\documentclass[a4paper,fleqn]{cas-sc}

\usepackage[numbers]{natbib}

\usepackage{amssymb}
\usepackage{float}
\usepackage{amsmath}
\usepackage[utf8]{inputenc}
\usepackage[ruled,vlined,linesnumbered]{algorithm2e}
\usepackage{graphicx}
\usepackage{multirow}
\usepackage{mathrsfs}
\usepackage{xcolor}
\usepackage{gensymb}
\usepackage{booktabs}
\usepackage{listings}
\usepackage{textcomp}
\usepackage{placeins}
\usepackage{lineno}
\usepackage{url}

\begin{document}
\setcounter{topnumber}{5}
\setcounter{bottomnumber}{5}
\setcounter{totalnumber}{10}
\renewcommand{\topfraction}{0.95}
\renewcommand{\bottomfraction}{0.95}
\renewcommand{\textfraction}{0.05}
\renewcommand{\floatpagefraction}{0.8}
\let\WriteBookmarks\relax
\def\textpagefraction{.001}

\shorttitle{Fringe Projection Based Vision Pipeline for Autonomous Hard Drive Disassembly}
\shortauthors{Balasubramaniam et al.}

\title[mode=title]{Fringe Projection Based Vision Pipeline for Autonomous Hard Drive Disassembly}

\author[1]{Badrinath Balasubramaniam}
\author[2]{Vignesh Suresh}
\author[3]{Benjamin Metcalf}
\author[3]{Beiwen Li}
\cormark[1]
\ead{beiwen.li@uga.edu}

\affiliation[1]{organization={School of Electrical and Computer Engineering, University of Georgia},
            city={Athens},
            postcode={30605},
            state={GA},
            country={USA}}

\affiliation[2]{organization={Alcon Research Laboratories},
            city={Fort Worth},
            postcode={76134},
            state={Texas},
            country={USA}}

\affiliation[3]{organization={School of Environmental, Civil, Agricultural and Mechanical Engineering, University of Georgia},
            city={Athens},
            postcode={30605},
            state={GA},
            country={USA}}

\cortext[1]{Corresponding author}

\begin{abstract}
The 62 million metric tons of e-waste generated in 2022, with the majority of the materials being unrecovered due to lack of infrastructure, represent a significant economic loss. Hard disk drives (HDDs) comprise a valuable and important e-waste stream, thereby necessitating robotic disassembly. Automating the disassembly of HDDs requires holistic 3D sensing, scene understanding, and fastener localization capabilities, however current methods for HDD e-waste recovery are fragmented, do not have robust 3D sensing capabilities, and do not focus on fastener localization.  We propose an autonomous vision pipeline which performs 3D sensing using a Fringe Projection Profilometry (FPP) based module, with selective triggering of a deep-learning-based depth completion module where FPP fails, and integrates this module with a lightweight, real-time instance segmentation network for scene understanding and critical component localization. Furthermore, since we utilize the same FPP camera-projector system which serves as input to both our depth sensing as well as component localization modules, our depth maps and derived 3D geometry are inherently pixel-wise aligned with the segmentation masks without the need for registration, providing an advantage over RGB-D perception systems common in industrial sensing.  We optimize both our trained depth completion and instance segmentation networks for deployment-oriented inference and quantify the inference speeds. The proposed system achieves a box mAP@50 of $0.960$ and mask mAP@50 of $0.957$ for instance segmentation, while the selected depth completion configuration with the Depth Anything V2 Base backbone achieves an RMSE of \(2.317\) mm and MAE of \(1.836\) mm; the Platter Facing learned inference stack achieved a combined latency of \(12.86\) ms and a throughput of \(77.7\) Frames Per Second (FPS) on the evaluation workstation. Finally, we adopt a sim-to-real transfer learning approach to augment our physical dataset. The proposed perception pipeline provides both high-fidelity semantic and spatial data which can be valuable for downstream robotic disassembly. To facilitate future research in this domain, the synthetic dataset developed for HDD instance segmentation will be made publicly available upon publication.

\end{abstract}


\maketitle

\section{Introduction}\label{sec1}
The amount of e-waste generated reached 62 million metric tons in 2022, with the economic value of the metals contained estimated at USD 91 billion. A majority of these materials go unrecovered and equate to billions of dollars of lost recyclable value annually \cite{balde2024global}. Furthermore, a large portion of this waste stream comprises hard disk drives (HDDs) which contain valuable materials such as gold, nickel, aluminum, palladium as well as the rare earth metal Neodymium, which stand to be recovered upon recycling  \cite{sabbaghi_global_2019,habib_tracking_2015,tanvar_characterization_2020}. However, recycling and material recovery rates are abysmally low due to a lack of recycling infrastructure in the industry \cite{liu_global_2023}. This, in turn, motivates the development of autonomous robotic disassembly systems to improve material recovery rates.

Recent research has explored various computer vision modalities to automate e-waste processing. Barman et al. \cite{barman_optimizing_2025} and Hussein et al. \cite{hussein_harnessing_2025} created a U-Net based framework that used hyperspectral images as input for their vision system that tackled E-waste recycling. Barman et al. \cite{barman_optimizing_2025} achieved printed circuit board (PCB) segmentation, identifying the integrated circuit, capacitor and connectors for E-waste recycling, while Hussein et al. \cite{hussein_harnessing_2025} used this for material classification; including copper, brass, aluminum, stainless steel, and white copper; on the Tecnalia WEEE dataset \cite{picon_analysis_2025}.  However, a hyperspectral imaging set up can be prohibitive in terms of cost, can be cumbersome to calibrate, and also suffer from slower scan speeds compared to RGB solutions. 

To address the limitations of hyperspectral imaging, researchers have optimized RGB-based deep learning frameworks for component localization. Mohsin et al. \cite{mohsin_heatmap_2025} proposed using ResNet-50 to classify capacitors, coils and diodes in PCBs, employing Grad-CAM \cite{selvaraju_grad-cam_2017} for weak localization. While efficient, this approach yields only rough class activation maps that may be insufficient for manipulation. 
Ahmed \& Sanam \cite{ahmed_exploration_2025} proposed a tag based sorting system utilizing YOLOv8n, for e-waste comprising phones, CPUs and batteries, which characterized the component's recyclability, toxicity, storage costs, material recovery and repairability. Jahanian et al. \cite{jahanian_see_2019} developed a multi-task learning framework that detects the PCB and associated small components in phones. However, the works surveyed are limited to PCB detection, which while valuable, is far from the full spectrum of components that can be recovered from complex electronics that end up discarded in landfills.  Liu et al. \cite{liu_raise_2026} demonstrated a fully automated system for smartphone disassembly using 2D images from an Intel RealSense D435i camera for localization. They employed YOLOv8 for object detection and instance segmentation to classify high and low value phone layers (separated by adaptive cutting), achieving 98.9\% sorting accuracy. This approach worked without depth sensing because the objects scanned had a small depth range, and the robotic system utilized vacuum grippers for pick and place operations, whereas high-fidelity 3D data is typically required for complex tasks such as fastener manipulation. While the above works excel at classification, sorting and surface-level detection, they do not possess the precision required for complex disassembly tasks such as fastener manipulation.

Multiple researchers have attempted to address this gap by developing fastener detection and localization techniques. Foo et al. \cite{foo_screw_2021} outlined a screw detection technique for crosshead fasteners (commonly used in Liquid Crystal Display (LCD) monitors), using a combination of deep learning and reasoning methods. A disadvantage of this technique is the hardcoded visual reasoning logic baked into the system. Although this helps the system achieve the desired accuracy, it comes at the cost of versatility and adaptability, with any possible samples that fall outside of these rules (either due to manufacturing differences or a damaged sample) incorrectly diagnosed.  Karbouj et al. \cite{karbouj_comparative_2024} conducted a comparative evaluation of one-stage (YOLOv5) and two-stage (Faster R-CNN) detectors for classifying screw heads on PC components, and demonstrated that one-stage detectors outperformed two-stage detectors in speed, precision, and recall. Indeed, the YOLO family has become the standard for real-time robotic e-waste disassembly, showing robust performance in isolating electronic boards \cite{puttero_automatic_2024} and localizing small fasteners \cite{brogan_deep_2021}. This is a motivating factor for us to use the more recent YOLOv11n in our own framework, which is a one-stage detector with only {\raise.17ex\hbox{$\scriptstyle\sim$}}2.9 million parameters.

While individual component detection and fastener detection are valuable, fine-scale disassembly of automated electronics requires accurate 3D spatial data for robot navigation.  Recent work has applied high dynamic range (HDR)-based fringe projection to acquire high-fidelity 3D geometry for e-waste assemblies such as desktop computers as a first step towards a perception system for downstream robotic disassembly \cite{zhao_precision_2025}. However, a complete disassembly perception system still requires this geometric sensing capability to be integrated with holistic scene understanding and fastener localization.
% However, this work still lacks an end to end pipeline comprised of holistic scene understanding, fastener localization, and 3D geometry acquisition, within the same framework, that is necessary for a complete disassembly perception system.

Yildiz et al. \cite{yildiz_visual_2020,yildiz_extended_2022} developed a complete vision system for automated hard drive disassembly, including 3D sensing, component localization, and fastener detection. They used Mask R-CNN for segmentation and a circular hough transform for screw detection. However, their approach has the following limitations: their pipeline is inaccurate at depth sensing for reflective regions, impeding fine-scale robotic disassembly. Furthermore, the mean screw detection precision of $80\%$ is far too unreliable, primarily because the authors rely on the circular hough transform for candidate generation \cite{yildiz_dcnn-based_2019}. Furthermore, Rojas et al. \cite{rojas_efficient_2023,rojas_deep_2022} improved upon Yildiz et al.'s localization by accounting for occlusion and varied angles, yet they did not address the fundamental limitations in depth sensing or the identification of specific disassembly fasteners.

 We have developed a low-cost, high-speed 3D vision system based on Fringe Projection Profilometry (FPP), a technique capable of kilohertz speeds and sub-millimeter accuracy suitable for industrial sensing \cite{li_high-speed_2022}. HDDs, however, present a severe optical challenge due to the coexistence of mirror-like platter regions, reflective metallic components, and absorptive surfaces within a compact assembly. We propose an FPP-based, logic-driven scanning protocol that performs scene understanding and critical component localization, including fasteners, and uses the detected HDD orientation to selectively trigger the Multi-Modal Depth Completion Network (MMDC-Net) (introduced in prior work \cite{balasubramaniam_application-driven_2026} as a depth completion module for optically challenging surfaces) when the mirror-like central platter \cite{noauthor_imagine_nodate} is present, thereby resolving reflective-region artifacts while maintaining efficiency on the back of the hard drive. An important advantage of our pipeline is the direct pixel-wise correspondence between the segmentation masks and reconstructed geometry, eliminating the need for an additional registration stage and simplifying downstream component localization. Our specific, itemized contributions are:

\begin{itemize}
    \item We propose a comprehensive vision pipeline for automated HDD disassembly by integrating high-speed FPP-based 3D sensing, MMDC-Net-based depth completion, and YOLOv11n-based segmentation, thereby enabling accurate component localization and dense reconstruction of optically challenging HDD regions.
    \item We demonstrate a sim-to-real workflow by training YOLOv11n on a synthetic dataset and achieving segmentation by fine-tuning on a smaller set of real-world HDD images, including critical disassembly components such as screws.
    \item We open-source our synthetic dataset comprising 3685 images designed specifically for HDD instance segmentation.
    \item We optimize the proposed pipeline for practical low-latency inference by evaluating Depth Anything V2 backbones to balance accuracy and speed, and by benchmarking FP16 Open Neural Network Exchange (ONNX)/TensorRT neural inference for the learned modules of the pipeline.
\end{itemize}

\begin{figure}[pos=htbp]
\centering
\includegraphics[width=0.8\textwidth]{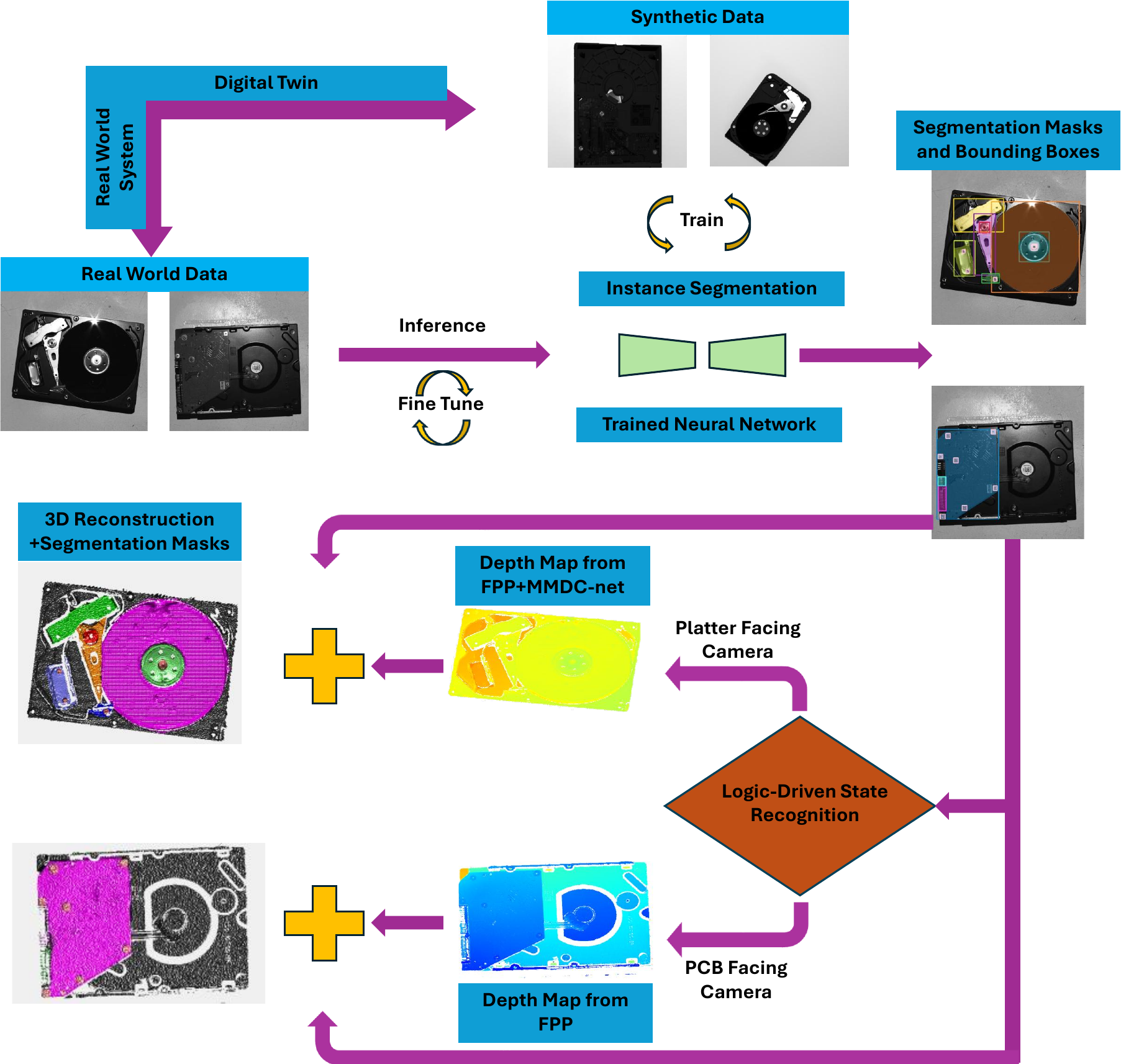}
\caption{Overall workflow of the proposed autonomous HDD disassembly vision pipeline. A calibrated digital twin is used to generate synthetic HDD images for pre-training the YOLOv11n instance segmentation model, which is subsequently fine-tuned on projector-illuminated real-world HDD images. During inference, the predicted masks are used both for component localization and for logic-driven state recognition. If the HDD is Platter Facing, the pipeline combines FPP with MMDC-Net to complete unreliable depth in reflective regions; if the HDD is PCB Facing, standard FPP is used directly. The final output is a semantically annotated 3D reconstruction for downstream robotic disassembly.}
\label{fig:overall_pipeline}
\end{figure}

Figure \ref{fig:overall_pipeline} outlines the complete algorithm.

\section{Methods}\label{sec2}
Autonomous disassembly requires 3D sensing, scene understanding and localization. We propose a vision pipeline where 2D semantic data guides the 3D profile extraction technique used. We start with the instance segmentation framework that localizes components, utilize its output to guide our logic-driven state recognition step, leverage the output of this step to guide our adaptive FPP system used for depth map computation.
\subsection{Instance Segmentation of HDD}
Accurate detection and localization of constituent components, including disassembly parts such as screws, can provide critical path planning information to the robot arm, and guide its positioning. To achieve this, we employ a deep-learning-based instance segmentation framework that processes 2D projector illuminated images to generate pixel-wise masks for HDD components, enabling both localization targeting and logic-driven state recognition described in the following subsections. The following subsection defines the component taxonomy for this task and the neural network architecture utilized for this.
\subsubsection{Segmentation Taxonomy}
\begin{figure}[pos=htbp]
\centering
\includegraphics[width=0.75\textwidth]{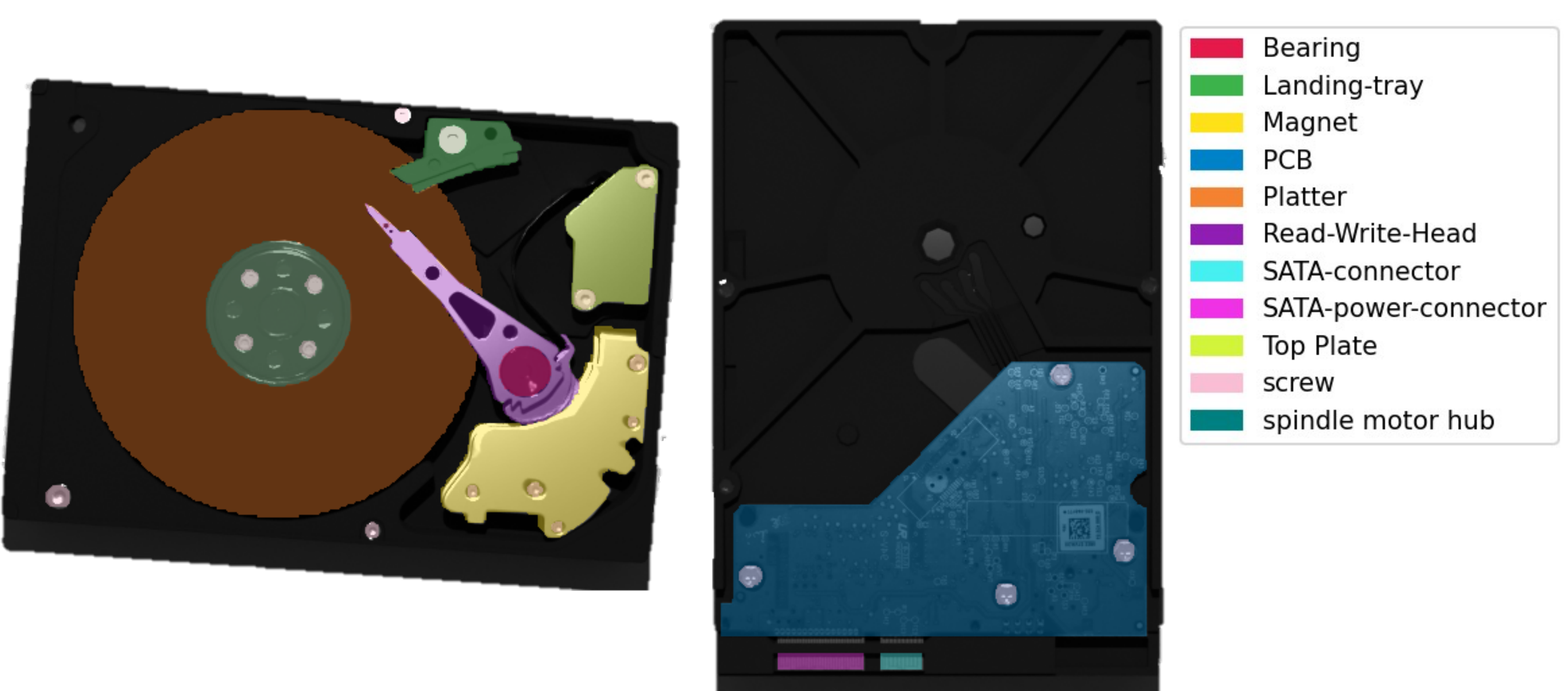}
% \caption{HDD Component Taxonomy }
\caption{Representative HDD component taxonomy used for instance segmentation. Left: Platter Facing view showing the internal mechanical components after removal of the top cover. Right: PCB Facing view showing the external electronics side of the drive. Colors denote the 11 component classes used in the proposed semantic segmentation framework, spanning mechanical and moving parts, electronics and interfaces, and fasteners.}
\label{fig:taxonomy}
\end{figure}

To facilitate a comprehensive disassembly process, we established a  taxonomy comprising 11 distinct HDD component classes. As detailed in Table \ref{tab:hdd_taxonomy} and pictorially depicted in Fig. \ref{fig:taxonomy}, these components are categorized into three functional groups:  \textit{Mechanical \& Moving}, \textit{Electronics \& Interfaces}, and \textit{Fasteners}.

 By explicitly distinguishing high-value components (such as PCBs containing precious metals, Neodymium magnets), fasteners, other moderate value metallic components (such as the bearing, read-write head, spindle, platter), connecting interfaces, and  
 miscellaneous modular components (e.g: landing tray), the vision system provides the semantic understanding required to formulate a disassembly sequence. 

\begin{table}[h!]
\centering
\caption{Taxonomy of Hard Disk Drive (HDD) components identified by the proposed semantic segmentation framework. The 11 classes are categorized by their functional role in the assembly.}
\label{tab:hdd_taxonomy}
\begin{tabular}{@{}ll@{}}
\toprule
\textbf{Functional Category} & \textbf{Component Class} \\ \midrule
% \multirow{1}{*}{\textbf{Structural \& Housing}} & Top Plate \\ \midrule
 
\multirow{6}{*}{\textbf{Mechanical \& Moving}} & Platter \\
 & Spindle Motor Hub \\
 & Top Plate \\ 
 & Read-Write-Head \\
 & Bearing \\
 & Landing Tray \\ \midrule
\multirow{4}{*}{\textbf{Electronics \& Interfaces}} & PCB \\
 & Magnet \\
 & SATA Connector \\
 & SATA Power Connector \\ \midrule
\textbf{Fasteners} & Screw \\ \bottomrule
\end{tabular}
 \end{table}

\subsubsection{Neural Network Architecture for Instance Segmentation}
We select YOLOv11-nano for instance segmentation. As established earlier, the YOLO family has been highly effective for e-waste localization; we utilize the nano variant ($\sim$2.9 M parameters) to maintain real-time performance while keeping the pipeline computationally lightweight.

\begin{figure}[pos=htbp]
\centering
\includegraphics[width=0.75\textwidth]{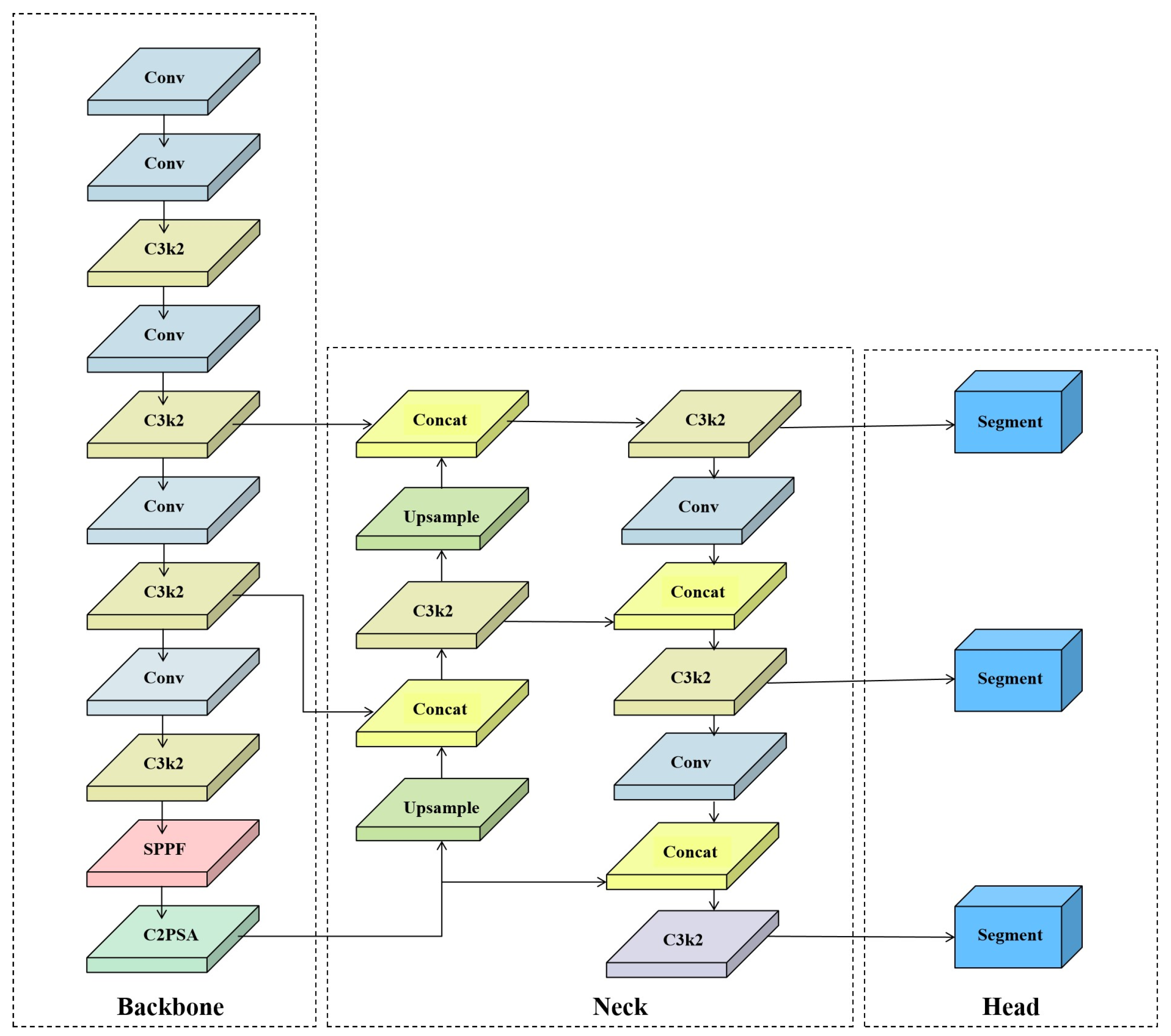}
\caption{Schematic of YOLOv11n-seg (Re-used from \cite{wang_non-contact_2025}) }
\label{fig:yolov11nseg_schematic}
\end{figure}

The structure of the YOLOv11 model \cite{khanam_yolov11_2024}, shown in Fig. \ref{fig:yolov11nseg_schematic}, comprises three major components: the backbone, the neck, and the head. The backbone is the primary feature extractor which creates multi scale feature maps from the input image. The neck is the intermediate component that aggregates and enhances feature representations across various scales. The head is the component that generates the final outputs, including bounding boxes, objectness scores, segmentation masks, and class labels.
This version introduces several novel components in comparison to its predecessor. For instance, it uses the C3k2 block throughout the architecture, replacing the C2f block from YOLOv8. This block is essentially a more computationally efficient version of the Cross Stage Partial Bottleneck and uses two smaller convolutions instead of a large one. It also introduces the C2PSA block in the backbone, which is a convolutional block with parallel spatial attention. This allows the model to focus on the regions of the image which may be important, and perform correlations which may be visually far apart in the image. The fast spatial pyramid pooling in the backbone, introduced originally in YOLOv5, has been retained. This feature helps aggregate features regardless of input size and increases the receptive field. The CBS blocks, used in the prediction head of the network for feature map refinement, have also been retained by YOLOv11 \cite{khanam_yolov11_2024}.

The nano variant (YOLOv11n) of the YOLOv11 is the fastest and most lightweight model in the YOLOv11 family. This makes it ideal for training when the amount of data is scarce since a large model can tend to overfit on insufficient training data, as well as for deployment in resource-constrained real-time inference settings. This, along with its inference speed, is the reason for our choice of this architectural variant. We leverage the model's pre-training on COCO-seg dataset and use these pre-trained weights as the starting points for all of our experiments. We use the segmentation masks and the bounding boxes of the output detected by our model for localization. We also use it for our logic-driven state recognition algorithm, described in the next subsection.

\subsection{Logic-Driven State Recognition}
To enable adaptive 3D scanning, the system must determine the HDD's state relative to the camera. We implement a State Recognition module that classifies the drive's state as either ``Platter Facing'' (front / face-up; cover removed, exposing internal components, facing camera) or ``PCB Facing'' (back / face-down; PCB on the rear facing camera). This classification relies on the semantic output provided by the instance segmentation network. Since we limit our analysis to 2.5 and 3.5 inch mechanical read-write based hard drives, each of which possesses an optically reflective central platter, we utilize the following rationale for this algorithm. If the Platter class is detected in the instance segmentation output, then our HDD orientation is characterized as ``Platter Facing''. Otherwise, our HDD orientation is characterized as ``PCB Facing''. This binary state classification dictates the subsequent 3D acquisition protocol, and is explained in more detail in the following subsection. 

\begin{equation}
    \text{State} =
    \begin{cases}
        \text{Platter Facing}, & \text{if } \exists \text{ mask}_{\text{platter}} \\
        \text{PCB Facing}, & \text{otherwise}
    \end{cases}
\end{equation}

\subsection{Adaptive 3D Profiling via FPP}
Following state recognition of the hard drive, the system initiates the 3D acquisition process. We utilize the FPP setup described in detail in the following subsection to generate the baseline depth map. However, to address the optical heterogeneity of hard drive components, the reconstruction pipeline adapts based on the detected state. For the PCB Facing state, the surface is predominantly matte and texture-rich. In this scenario, standard phase-shifting and triangulation are sufficient to generate a high-fidelity point cloud with sub-millimeter accuracy. Conversely, the Platter Facing state exposes the mirror-like central platter, and other metallic and optically challenging components that cause saturation and phase unwrapping failures in structured light systems. Upon detecting this state, the pipeline selectively activates the MMDC-Net. This module completes the depth map computed from FPP, effectively resolving artifacts where the geometric sensor is unreliable. The architectural details and training methodology for MMDC-Net are established in our prior work \cite{balasubramaniam_application-driven_2026}.

\subsubsection{Fringe Projection Profilometry}
\begin{figure}[pos=htbp]
\centering
\includegraphics[width=0.75\textwidth]{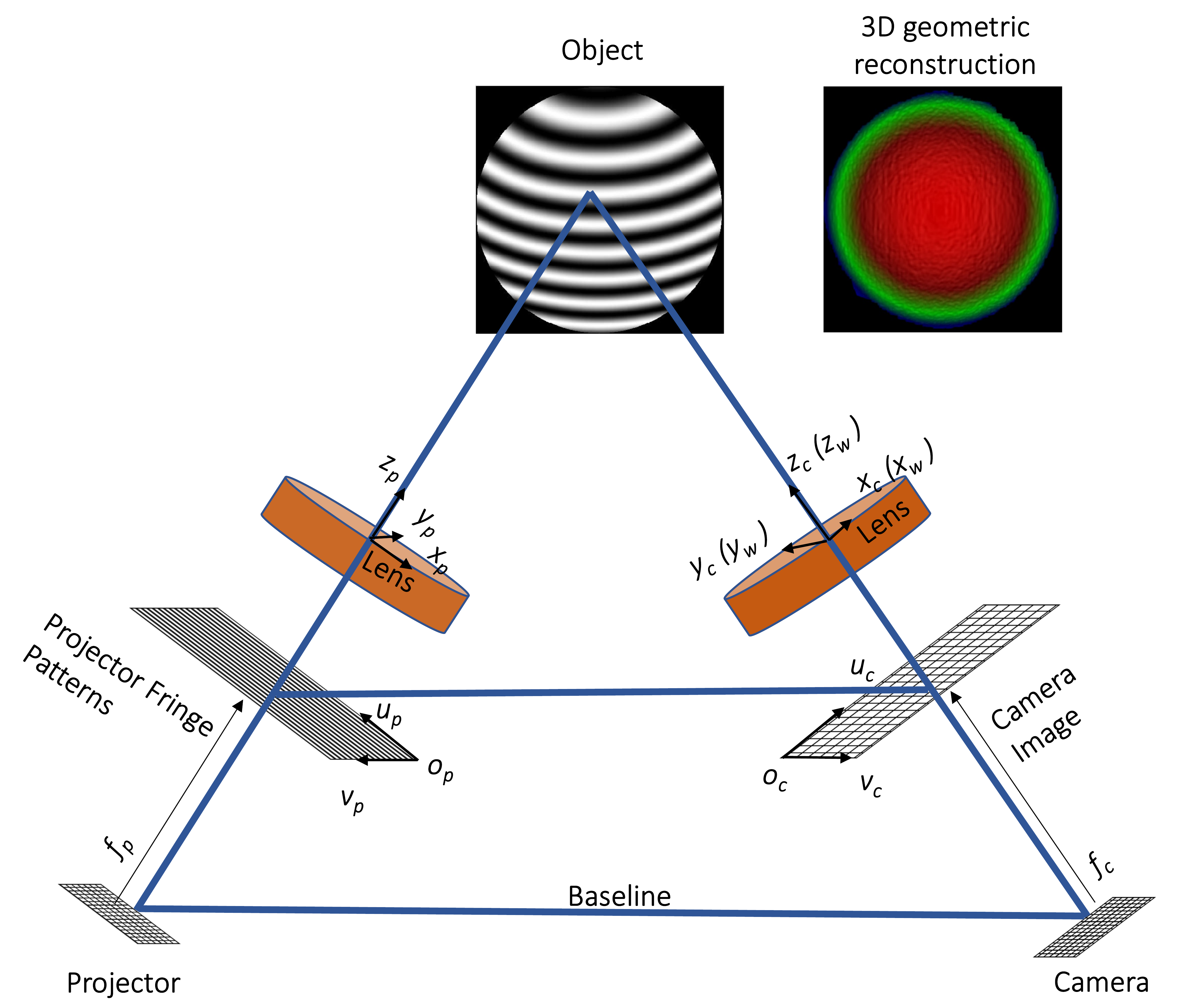}
\caption{Schematic of FPP (Adapted from \cite{naqvi_four-dimensional_2025}) }
\label{fig:FPP_schematic}
\end{figure}

FPP is a 3D vision technique that involves the use of a camera and a projector. The projector projects black and white fringe patterns on the object. These fringe patterns then deform according to the shape of the object, encoding the depth of the object at each pixel. An image of our real-world system is outlined in Fig. \ref{fig:FPP_schematic}. Fringe projection profilometry has many advantages over other 3D scanning technologies. The ability to acquire high-speed scans (kilohertz-rate acquisition using binary pattern projection with optical defocusing), while achieving sub millimeter accuracy and performing simultaneous whole area scanning, contributes to our choice of this technique over stereo vision, which can rely on surface texture for correspondence matching and struggle with feature sparse regions; laser scanning, which can be much slower and is done line by line; and time of flight, which is not as accurate as FPP and can be influenced to a greater degree by ambient light.

The intensity \(I\) of the \(n^{\text{th}}\) phase-shifted image can be written as
\begin{equation}
I_n(x,y) = I'(x,y) + I''(x,y)\cos\!\big(\phi(x,y) + \delta_n\big), \qquad n = 1,2,3,\ldots,18
\end{equation}
where \(I'(x,y)\) is the average intensity and \(I''(x,y)\) is the fringe modulation. $n$ ranges from 1 through 18 due to utilizing the 18-step phase shifting algorithm. The wrapped phase is then computed using the standard phase-shifting formulation:
\begin{equation}
\phi(x,y) = -\arctan\!\left(\frac{\sum_{n=1}^{18} I_n \sin \delta_n}{\sum_{n=1}^{18} I_n \cos \delta_n}\right).
\end{equation}

We unwrap this phase using the graycoding technique \cite{zhang_high-speed_2018} to generate the absolute phase. Using the system calibration parameters derived from Li et al. \cite{li_novel_2014}, we map the absolute phase and camera coordinates $(u_c, v_c)$ to the world coordinates $(x_w, y_w, z_w)$ via triangulation, generating the dense 3D point cloud required for robotic maneuvering. This 3D point cloud has high fidelity for the PCB Facing HDDs. The depth completion module, used in conjunction with FPP for the more complex optical surfaces encountered in the Platter Facing HDD case, is explained in the next subsection.
\subsubsection{MMDC-Net-Based Depth Completion for Optically Challenging Surfaces}
For Platter Facing HDDs, severe specular reflections and underexposed regions can corrupt the raw FPP reconstruction and produce sparse or unreliable depth. To address this, the proposed pipeline selectively invokes MMDC-Net, which fuses the sparse FPP depth map, the projector-illuminated grayscale image, and a relative depth prior from Depth Anything V2 to predict dense depth in unreliable regions. The predicted depth is then combined with the reliable FPP measurements to generate a complete depth map. Since the architecture and training procedure of MMDC-Net are detailed previously in \cite{balasubramaniam_application-driven_2026}, only its role within the present adaptive pipeline is described here.
\section{Experiments}\label{sec3}
In this section, we detail the experimental protocols, beginning with the set up and calibration of the physical FPP scanning system. We outline our computational hardware and subsequently describe the development of the digital twin used for synthetic training data generation, followed by the specific hyperparameters and data augmentation strategies employed to train neural networks for instance segmentation.

\subsection{Real-World Imaging System and Computational Hardware Set-Up}
\begin{figure}[pos=htbp]
\centering
\includegraphics[width=0.5\textwidth]{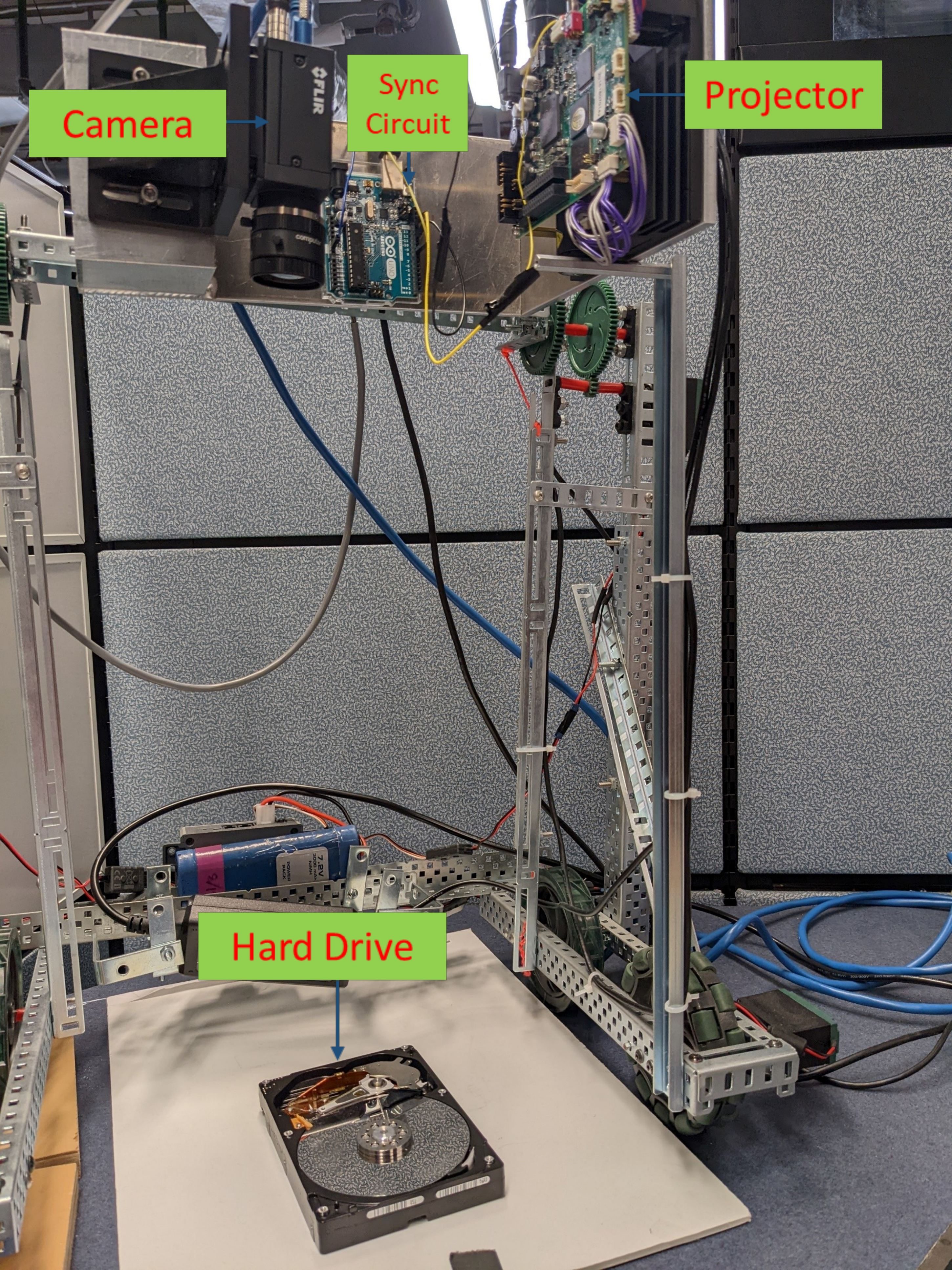}
\caption{Real-World Hardware Set Up (Adapted from \cite{balasubramaniam_3d_2023}) }
\label{fig:hardware_schematic}
\end{figure}

The real-world imaging setup, shown in Fig. \ref{fig:hardware_schematic}, consists of a DLP 4500 Lightcrafter projector for pattern projection, a complementary metal-oxide-semiconductor (CMOS) (model: FLIR Grasshopper3 GS3-U3-23S6M-C) for image acquisition, and an Arduino Uno for camera-projector synchronization.

\textbf{Calibration}: System calibration is performed using a $5 \times 9$ circle grid target using the technique proposed in Li et al \cite{li_novel_2014}. It involves rotating the circle board through 18 different poses and capturing 52 images of horizontal and vertical fringe patterns. The horizontal and vertical phase maps computed are then used in conjunction with OpenCV's camera calibration toolbox to compute the projector circle centers. This is then used to compute the projector intrinsic matrix. The world coordinate system is assumed to be aligned with the camera coordinate system. The projector extrinsic matrix is then computed using the OpenCV stereo calibration toolbox. A detailed account of the steps taken for our procedure is provided in Balasubramaniam et al. \cite{balasubramaniam_single_2023}.

\textbf{Acquisition Protocol}: To ensure high fidelity depth maps and high accuracy, we configured the FPP system to utilize an 18-step phase-shifting algorithm ($N=18$). We employed binary defocusing to project 1-bit quasi-sinusoidal patterns, enabling high-speed projection rates \cite{li_high-speed_2022}. We used the graycoding technique for phase unwrapping, projecting a total of 24 fringe patterned images + 1 projector illuminated image per scan.
We have explained the real-world data collection process for the 3D data (to train MMDC-Net) in \cite{balasubramaniam_application-driven_2026}. For the instance segmentation data generation, we used the projector illuminated grayscale images of 40 different hard drives as the input,  and utilized Roboflow \cite{roboflow2026} to label these images. 

\textbf{Computational Hardware Setup}: Our computational hardware is identical to \cite{balasubramaniam_application-driven_2026}, with all data generation, neural network training and evaluation experiments performed on a machine with an NVIDIA RTX 5000 Ada GPU (32 GB VRAM), an Intel Xeon w5-2465X processor (16 cores, 4.7 GHz max clock speed), and 128 GB of system RAM.
\subsection{Digital Twin Set-Up and Synthetic Data Generation}
We use Zheng et al.'s \cite{zheng_fringe_2020} digital twin developed in Blender for an FPP system. We adapted the digital twin for our application following the same procedure as in Balasubramaniam et al. \cite{balasubramaniam_application-driven_2026,balasubramaniam_single_2023}. The synthetic data generation process to train our depth completion neural network, MMDC-Net, has already been explained in \cite{balasubramaniam_application-driven_2026}. We shall hence restrict ourselves to the synthetic data generation process for instance segmentation.
\subsubsection{Synthetic Data Generation- Algorithm for Automatic Component Mask Extraction with Occlusion Handling}
\begin{figure}[pos=htbp]
\centering
\includegraphics[width=1\textwidth]{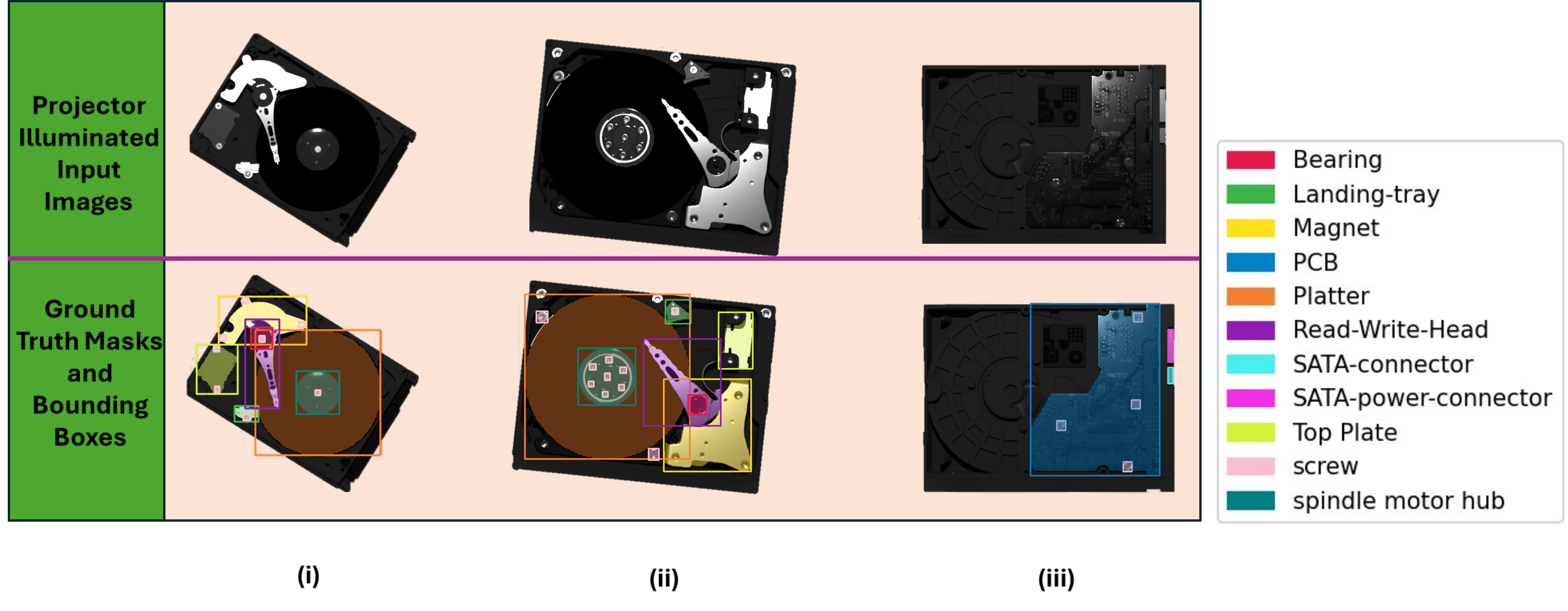}
\caption{Synthetic Data Input and Ground Truth Pairs with (i) Sample 2.5 inch HDD (front), (ii) Sample 3.5 inch HDD (front), and (iii) Sample 3.5 inch HDD (back)  }
\label{fig:synthetic_data}
\end{figure}
We utilize 14 CAD models sourced from SketchFab \cite{noauthor_sketchfab_nodate} and GrabCAD \cite{noauthor_grabcad_nodate}. The projector from our digital twin illuminates the object with white light. We rotate each CAD model through multiple orientations. At each orientation, we capture a projector illuminated grayscale image of the hard drive, as well as binary images containing segmentation masks of each component. 
\begin{algorithm}[h] 
\small % Keeps the text compact and professional
\caption{Synthetic Hard Drive Dataset Generation with Semantic Segmentation Masks}
\label{alg:hdd_dataset_generation}

\SetKw{KwBy}{by}
\SetKwComment{Comment}{// }{}

% Updated Data and Result to reflect single image per rotation
\KwData{3D hard drive CAD model, single fringe pattern (white light) $P$, material definitions $M = \{m_1 \ldots m_K\}$}
\KwResult{Rendered image $I_{\theta}$ and segmentation masks $S_{\theta,k}$ for each rotation $\theta$}

\SetKwFunction{RenderMasks}{RenderMaterialMasks}
\SetKwFunction{RenderFringe}{RenderFringePattern}
\SetKwFunction{ModifyMaterials}{RandomizeMaterialProps}
\SetKwFunction{SetSpotlight}{SetSpotlightTexture}
\SetKwFunction{ClearCache}{ClearPatternCache}

\tcp{Initialization}
Load configuration: pattern directory, rotation parameters\;
\ClearCache{}\; 

\tcp{Main rendering loop}
\For{$\theta = 0$ \KwTo $\theta_{max}$ \KwBy $\Delta\theta$}{
    
    \tcp{Apply rotation to 3D model}
    $R_{\text{model}} \leftarrow R_{\text{model}} \cdot R_z(\Delta\theta)$\;
    
    \tcp{Randomize material properties for domain variation}
    \ModifyMaterials{$M$, roughness $\in [r_{min}, r_{max}]$,}
    \Indp \Indp brightness $\in [v_{min}, v_{max}]$\; \Indm \Indm
    
    \tcp{Set neutral white lighting for mask generation}
    \SetSpotlight{$P_{\text{white}}$}\;
    Update scene lighting\;
    
    \tcp{Generate semantic segmentation masks}
    Enable Material Index render pass\;
    \For{$k = 1$ \KwTo $|M|$}{
        Assign material index: $m_k.\text{pass\_index} \leftarrow k$\;
    }
    
    \tcp{Render individual material masks}
    \For{$k = 1$ \KwTo $|M|$}{
        Configure ID Mask compositor node with index $k$\;
        $S_{\theta,k} \leftarrow$ \RenderMasks{$\theta, k$}\;
        Save: $S_{\theta,k} \rightarrow$ \texttt{masks/mat\_$k$\_$\theta$.png}\;
    }
    
    \tcp{Render projector illuminated grayscale image}
    Disable compositing\;
    \ClearCache{}\;
    \SetSpotlight{$P$}\;
    $I_{\theta} \leftarrow$ \RenderFringe{$\theta$}\;
    Save: $I_{\theta} \rightarrow$ \texttt{images/$\theta$.png}\;
}

\KwRet{Dataset $\mathcal{D} = \{(I_{\theta}, \{S_{\theta,k}\}_{k=1}^{|M|})\}_{\theta}$}

\end{algorithm}
 Following this process, we generate 3685 different image-mask pairs. Following the generation of these raw binary masks, we implemented an automated post-processing pipeline to convert them into the normalized polygon format required for training the YOLOv11n-seg model. This conversion process is divided into two distinct workflows based on component multiplicity:

For unique structural and electronic components (e.g., the platter, PCB, read-write head), where a maximum of one instance exists per HDD, we apply standard topological structural analysis to extract the external contours of the binary mask. The coordinates of the largest continuous contour are then normalized relative to the image dimensions to generate the standard YOLO segmentation label.

Conversely, fasteners such as screws present a distinct challenge. Because multiple screws often exist on a single hard drive, the initial render produces a combined binary mask containing multiple instances that must be separated to train an instance segmentation model. In our experiments, conventional morphology- and connected-component-based separation methods were not sufficiently robust across all orientations, particularly when nearby screw masks merged into a single connected region. We therefore use (Segment Anything Model 3) SAM 3 \cite{carion2025sam} as an offline instance-separation step during dataset generation. First, all screw masks for a given orientation are merged into a single binary image. We then prompt SAM 3 to identify discrete "white circles" within this merged mask, enabling it to separate independent instance masks for each screw. The external contours of these SAM-generated masks are subsequently extracted and normalized, successfully yielding separate YOLO bounding polygons for every screw in the scene. The input and corresponding ground truth YOLO masks and bounding box pairs are depicted in Fig. \ref{fig:synthetic_data}. The synthetic HDD instance segmentation dataset will be made publicly available upon publication, with access instructions and download links provided at the repository: \url{https://github.com/badri999/HDD-Segmentation-Synthetic-Data}.

\subsubsection{Synthetic Training- Hyperparameters and Data Augmentation}

\label{subsec:synthetic training}
We train our YOLOv11n-seg model (pre-trained on the COCO-seg) on the synthetic dataset consisting of 3685 image-mask pairs, each of size $512 \times 512\times 3$. The full set of hyperparameters used for training the model are outlined in Table \ref{tab:rw_hyperparams}.
\subsection{Real-World Data Acquisition}
Using the experimental setup in Figure \ref{fig:hardware_schematic}, we image 40 different hard drives. The 3D data acquisition and dataset composition is described in our previous paper \cite{balasubramaniam_application-driven_2026}. For the 2D data necessary for scene understanding, we utilize the projector to project white light, and acquire fully illuminated grayscale images of the hard drive. We then annotate these images using Roboflow \cite{roboflow2026} and increase the size of the dataset by rotating each image and its associated labels in four increments of 90\degree each.

\subsubsection{Real-World Fine-Tuning: Hyperparameters and Data Augmentation}
\label{subsec:real-world training}
We fine-tune our YOLOv11n-seg model (pre-trained on the synthetic dataset) on the real-world dataset consisting of 160 total annotated projector illuminated grayscale images, each of size $512 \times 512\times 3$. The full set of hyperparameters used for training the model are outlined in Table \ref{tab:rw_hyperparams}.

\begin{table}[htbp]
\centering
\caption{Hyperparameters used for training the YOLOv11n-seg instance segmentation model}
\label{tab:rw_hyperparams}
% \footnotesize % Reduces the font size
\renewcommand{\arraystretch}{1} % Reduces the vertical space between rows
\begin{tabular}{@{}ll@{}}
\toprule
\textbf{Hyperparameter} & \textbf{Value} \\ \midrule
\multicolumn{2}{c}{\textit{General Training}} \\ \midrule
Model  & YOLOv11n-seg\\
Max. Epochs & 300 \\
Batch Size & 16 \\
Precision & Mixed (AMP) \\
Optimizer & AdamW \\
Initial Learning Rate ($lr0$) & $5 \times 10^{-4}$ \\
Momentum & 0.9 \\
Weight Decay & $5 \times 10^{-4}$ \\ \midrule
\multicolumn{2}{c}{\textit{Geometric Augmentation}} \\ \midrule
Mosaic & 1.0 (100\%) \\
Flip Left-Right (\texttt{fliplr}) & 0.5 \\
Flip Up-Down (\texttt{flipud}) & 0.5 \\
Scale & 0.5 \\
Translation & $\pm 0.1$ \\ \midrule
\multicolumn{2}{c}{\textit{Photometric Augmentation (Albumentations)}} \\ \midrule
Blur / Median Blur & $p=0.01$ \\
Grayscale (\texttt{ToGray}) & $p=0.01$ \\
CLAHE & $p=0.01$ \\
HSV-H (Hue) & 0.015 \\
HSV-S (Saturation) & 0.7 \\
HSV-V (Value) & 0.4 \\ \bottomrule
\end{tabular}
\end{table}
\section{Results and Discussion}\label{sec11}

\subsection{Instance Segmentation of HDD Components}
\begin{figure}[pos=htbp]
\centering
\includegraphics[width=1\textwidth]{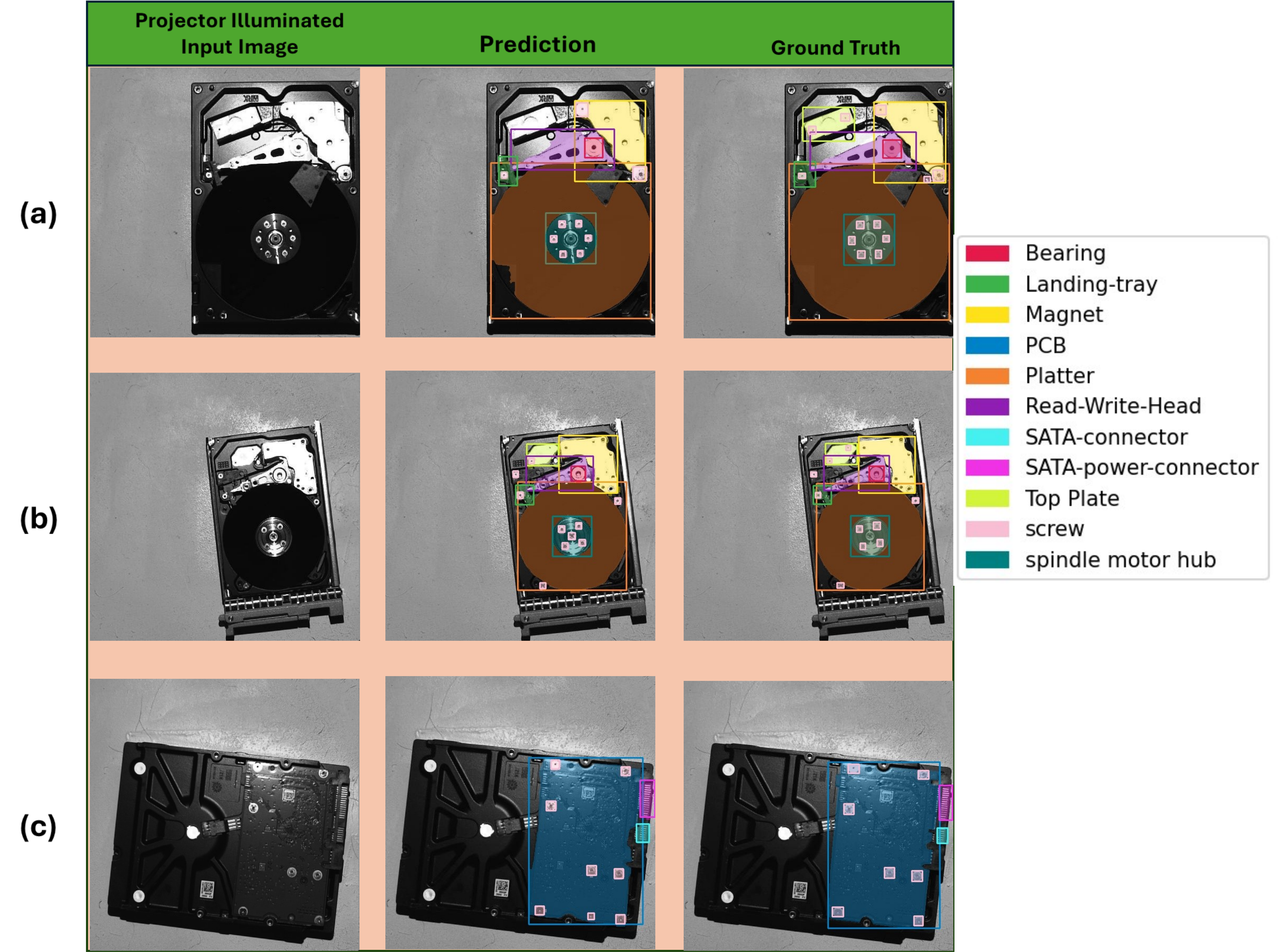}
% \caption{Instance Segmentation Sample Results on Hard Drives in the Test Set}
\caption{Qualitative instance segmentation results on representative unseen HDD test samples. Rows (a)–(c) show three different HDD examples spanning both Platter Facing and PCB Facing orientations. Columns show the projector-illuminated input image, the YOLOv11n-seg prediction, and the corresponding ground-truth annotation. The results illustrate accurate localization and boundary delineation of both macro-components and small fasteners across varying viewpoints and drive states.}
\label{fig:segmentation_results}
\end{figure}

In this section, we evaluate the performance of the YOLOv11n-seg architecture in localizing and segmenting the constituent components of the hard disk drive. Accurate instance segmentation serves as the foundational step of our pipeline, as it not only performs localization of critical components, but also provides the semantic context required for our logic-driven 3D scanning protocol. Qualitative results of the model's inference on unseen test samples are illustrated in Fig. \ref{fig:segmentation_results}, demonstrating robust boundary delineation across both Platter Facing and PCB Facing orientations.

We begin by pre-training the YOLOv11n-seg model on our generated synthetic data consisting of projector illuminated hard drive and ground truth component mask label pairs. We fine-tune this trained model on real-world data, and in order to ensure that every sample is tested once, we utilize k-fold cross validation, separating the data into eight folds, with five samples in each fold. In order to ensure that the network does not see the same sample at a different orientation in multiple folds, we separate the original 40 samples into 8 different folds before augmenting it with four rotations in increments of 90$\degree$.  The results achieved on real-world data are shown in Table \ref{tab:segmentation_results}, where we display the mean and standard deviation values of segmentation metrics across folds. We make the following observations:
\begin{itemize}
     \item \textbf{Overall Pipeline Metrics:} The overall Box mAP@50, Mask mAP@50, Box mAP@50--95, and Mask mAP@50--95 for our pipeline are high, with values of $0.960$, $0.957$, $0.809$, and $0.724$ respectively. %Furthermore, this is a significant improvement over the results achieved by Yildiz et al.\cite{yildiz_visual_2020}, who report an overall Mask mAP@50 of 0.767 and a Mask mAP@50-95 of 0.562 for their pipeline, which utilized Mobile-RCNN.

    \item \textbf{High and Medium Value Components:} High value components, including the \texttt{Magnet}  and \texttt{PCB}, and some mid-value metallic components, including  the \texttt{Platter}, \texttt{Spindle Motor Hub}, and  \texttt{Read-Write-Head}, consistently saturated the $0.995$ upper threshold across all K-folds, indicating high geometric reliability regardless of the training pipeline.
    
    \item \textbf{Critical Fasteners (Screws):} The localization of \texttt{Screws} is also exceptionally high, with a Box AP@50 of $0.929$, and a Mask AP@50 score of $0.910$. By priming the network with synthetic versions of these small components under varying lighting conditions, the model successfully detects and segments screws with high accuracy. %This is again a significant improvement over \cite{yildiz_visual_2020}, where the authors use a circular hough transform to generate candidates, resulting in an overall precision of $0.80$.

\end{itemize}

Furthermore, our instance segmentation pipeline compares favorably with prior HDD disassembly perception literature. Yildiz et al. \cite{yildiz_visual_2020} reported an aggregate $AP_{0.5}$ of $0.767$ using bounding-box-based evaluation, whereas our sim-to-real pipeline achieves a Box mAP@50 of $0.960$ and a Mask mAP@50 of $0.957$. In addition, while Yildiz et al. \cite{yildiz_visual_2020} relied on circular Hough transform candidate generation and downstream classification for screw detection, yielding an average precision of $80\%$, our unified architecture directly detects and segments screws with a Box AP@50 of $0.929$ and a Mask AP@50 of $0.910$. These results provide stronger localization fidelity for fasteners and pixel-level component boundaries, both of which are important for robotic disassembly.

Rojas et al. \cite{rojas_efficient_2023, rojas_deep_2022} reported an overall Mask AP@50--95 of $0.722$ using UNINEXT \cite{lin_uninext_2023}. Our pipeline achieves a comparable but slightly higher Mask mAP@50--95 of $0.724$ while using a sim-to-real workflow in which the majority of the training data are synthetic. In addition, our taxonomy includes fasteners, which are essential for robotic disassembly planning. Taken together, these results validate the instance segmentation portion of our autonomous hard drive disassembly pipeline.
\begin{table}[htbp]
\centering
\caption{Class-wise instance segmentation and object detection performance of the proposed Sim-to-Real pipeline (YOLOv11n-seg pre-trained on synthetic data and fine-tuned on real-world data). Metrics reported are Object Detection (Box AP@50) and Instance Segmentation (Mask AP@50).}
\label{tab:segmentation_results}
\resizebox{0.5\textwidth}{!}{%
\begin{tabular}{@{}lcc@{}}
\toprule
\textbf{Component Class} & \textbf{Box AP@50} & \textbf{Mask AP@50} \\ \midrule
% \multicolumn{3}{c}{\textbf{Structural \& Housing}} \\ \midrule

\multicolumn{3}{c}{\textbf{Mechanical \& Moving}} \\ \midrule
Platter & $0.995$ & $0.995$ \\
Top Plate & $0.975 \pm 0.047$ & $0.970 \pm 0.061$ \\
Spindle Motor Hub & $0.995$ & $0.995$ \\
Read-Write-Head & $0.995$ & $0.995$ \\
Bearing & $0.995$ & $0.995$ \\
Landing Tray & $0.934 \pm 0.157$ & $0.922 \pm 0.159$ \\
\multicolumn{3}{c}{\textbf{Electronics \& Interfaces}} \\ \midrule
PCB & $0.995$ & $0.995$ \\
Magnet & $0.995$ & $0.995$ \\
SATA Connector & $0.840 \pm 0.361$ & $0.866 \pm 0.294$ \\
SATA Power Connector & $0.888 \pm 0.283$ & $0.880 \pm 0.305$ \\
\multicolumn{3}{c}{\textbf{Fasteners}} \\ \midrule
Screw & $0.929 \pm 0.049$ & $0.910 \pm 0.062$ \\
\midrule
\multicolumn{3}{c}{\textbf{Overall Pipeline Performance}} \\ \midrule
Overall mAP@50 & $0.960 \pm 0.056$ & $0.957 \pm 0.053$ \\
Overall mAP@50--95 & $0.809 \pm 0.056$ & $0.724 \pm 0.062$ \\
\bottomrule
\end{tabular}%
}
\end{table}
\subsection{Logic-Driven State Recognition}\label{sec:logic}
In this subsection, we present the results of the Logic-Driven State Recognition step of our algorithm. The system evaluates the binary presence of the \texttt{Platter} class in the 2D segmentation mask output to classify the drive as Platter Facing.  The segmentation model achieved 100\% precision and 100\% recall for platter presence on the evaluated folds, correctly identifying every exposed platter and producing no false platter detections on PCB Facing images. This ensures that the deep-learning-based depth completion module correctly selectively triggers.

\subsection{3D Profile Extraction of HDD Components}
In this section, we evaluate the quality of the 3D point clouds generated by our proposed vision pipeline. We assess the geometric reconstruction performance across both the matte, PCB Facing orientation of the hard drive and the optically challenging, Platter Facing orientation. Given the scope of this work, quantitative analysis is focused on the reflective-region completion pathway and deployment-oriented trade-offs, while the PCB Facing FPP regime is treated as a qualitative baseline operating condition.

\subsubsection{PCB Facing HDD Reconstruction using Baseline FPP}
The back (PCB Facing) orientation of the hard disk drive predominantly exposes the PCB and various structural fasteners. Because these components do not exhibit large regions of high reflectivity, they present a surface that is conducive to standard structured light scanning. Since the PCB Facing HDD interior contains comparatively few highly specular regions, it is treated as a baseline operating condition for FPP and is evaluated qualitatively. Quantitative analysis is instead focused on the Platter Facing case, where strong platter reflections create the primary reconstruction challenge addressed in this work.

\begin{figure}[pos=htbp]
\centering
\includegraphics[width=0.75\textwidth]{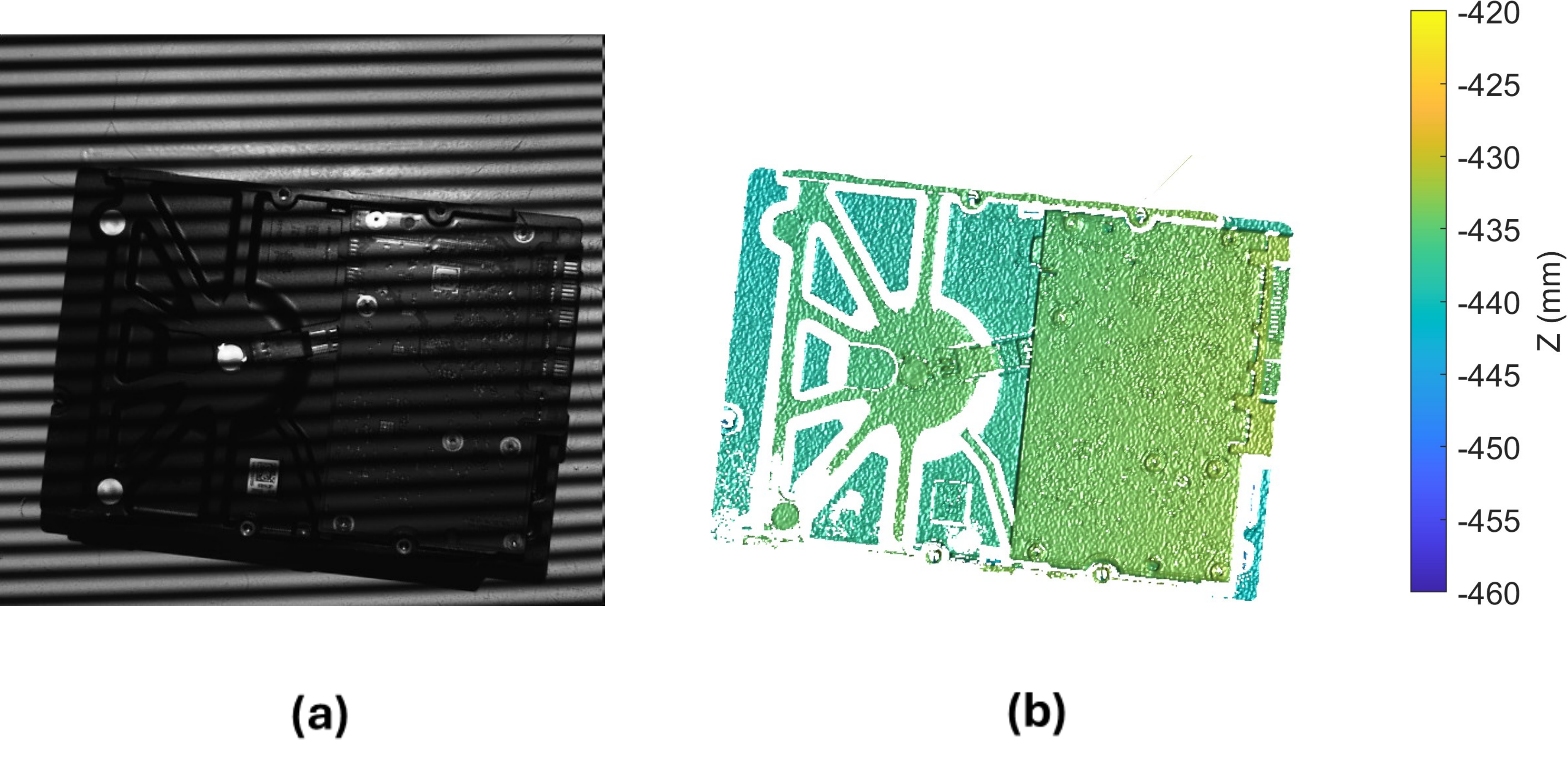}
% \caption{FPP performed on the back of the HDD }
\caption{Baseline FPP reconstruction for a PCB Facing HDD. (a) Representative fringe-pattern image acquired from the PCB Facing side of the drive. (b) Corresponding 3D reconstruction obtained using standard FPP, with color indicating depth $z$ in millimeters. Because the exposed PCB side is largely matte and contains few highly specular regions, baseline FPP is sufficient to recover the 3D profiles of the PCB and associated screws, aside from minor shadow-induced gaps.}
\label{fig:FPP_back_hdd}
\end{figure}

We demonstrate the baseline FPP system on the back of the HDD in Fig. \ref{fig:FPP_back_hdd}. We observe that the system achieves a high-fidelity 3D reconstruction without the need for physical surface treatments or deep-learning-based depth completion. Excluding minor data loss caused by occlusion-induced shadows, the resulting geometric reconstruction has high fidelity. This baseline FPP output is sufficient to reliably extract the 3D profiles of the PCB and its associated screws for downstream robotic manipulation.
\subsubsection{Platter Facing HDD Reconstruction with Selective MMDC-Net Depth Completion}\label{sec:results_hdd_faceup}
\begin{figure}[pos=htbp]
\centering
\includegraphics[width=0.75\textwidth]{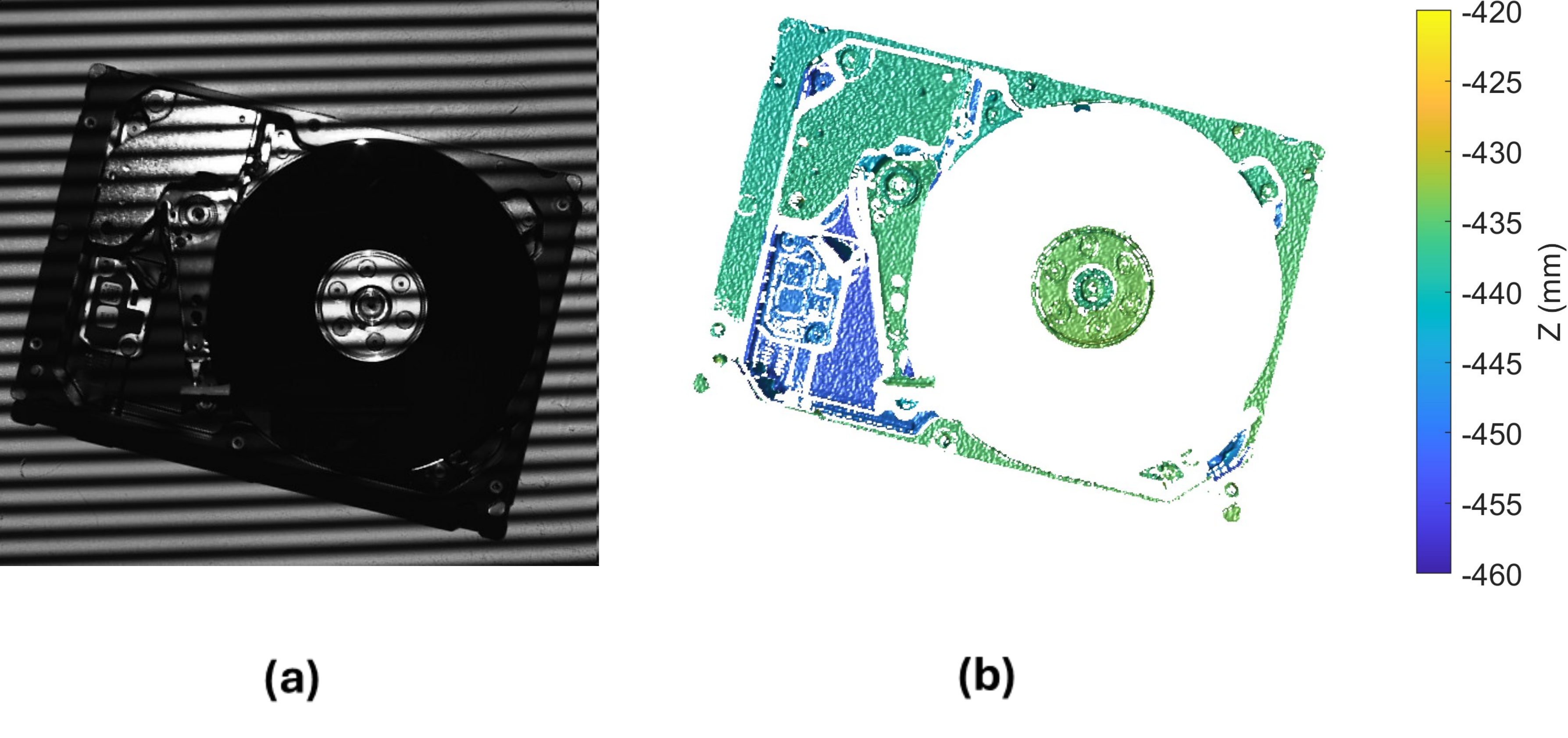}
% \caption{FPP performed on the back of the HDD }
\caption{Baseline FPP reconstruction for a Platter Facing HDD. (a) Fringe-pattern image acquired from the Platter Facing side of the drive. (b) Corresponding 3D reconstruction obtained using standard FPP, with color indicating depth z in millimeters. Strong specular reflections from the platter and reflective metallic regions lead to sparse reconstruction and missing depth, motivating the use of MMDC-Net in the proposed pipeline.}
\label{fig:FPP_front_sparse}
\end{figure}

The front (Platter Facing) orientation of the HDD presents a significantly more challenging optical environment due to the exposed, mirror-like central platter. When scanned using standard FPP, highly specular and reflective regions result in severe artifacts and missing data (holes) in the generated depth map. A representative sparse reconstruction obtained by baseline FPP on a Platter Facing HDD is shown in Fig. \ref{fig:FPP_front_sparse}.

\begin{figure}[pos=htbp]
\centering
\includegraphics[width=0.75\textwidth]{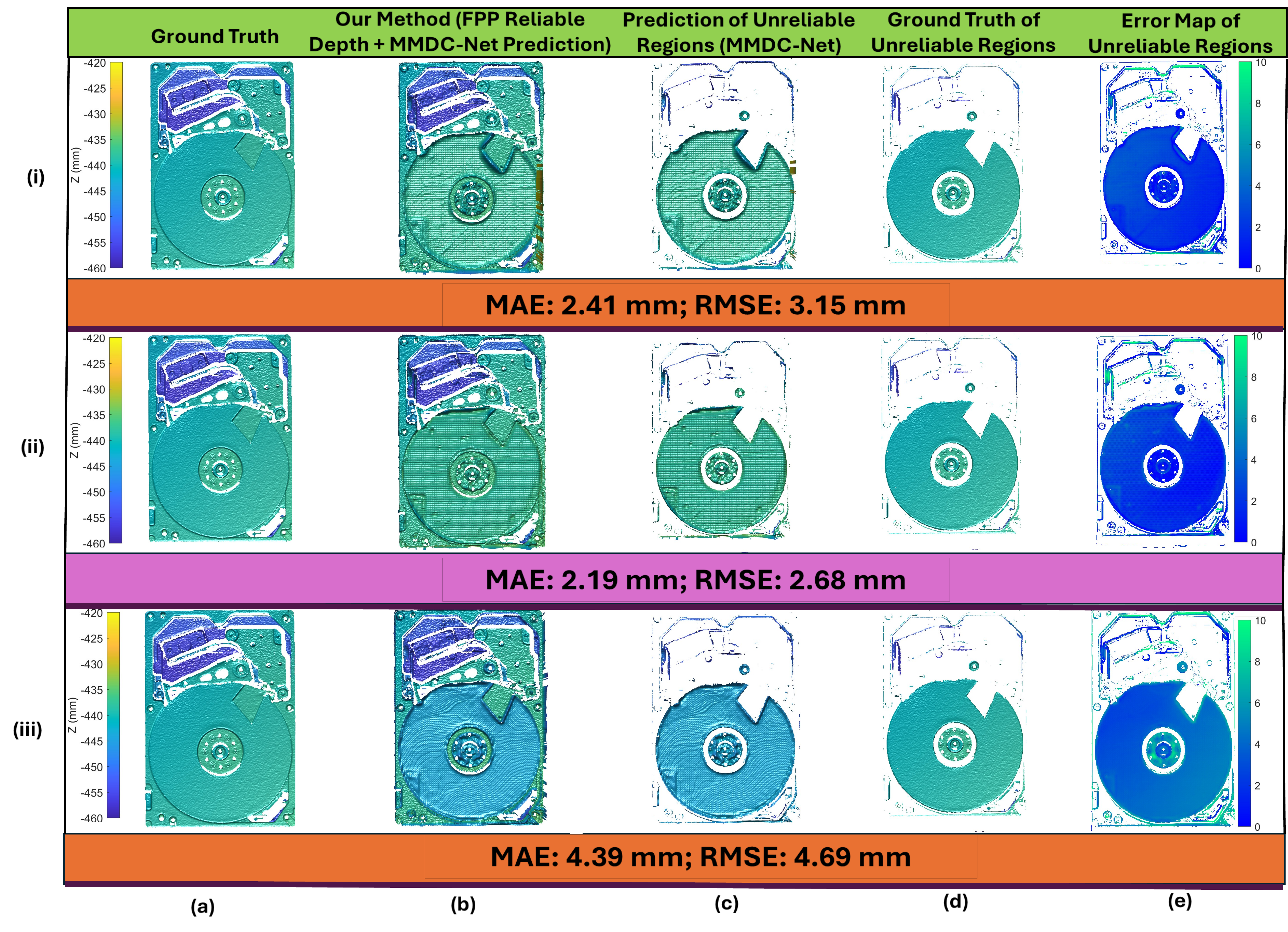}
% \caption{Comparison of MMDC-Net Trained on varying Depth Anything V2 Backbones }
\caption{Qualitative comparison of MMDC-Net variants trained with different Depth Anything V2 backbones on a representative Platter Facing HDD. Rows (i)–(iii) correspond to Depth Anything V2 Large, Base, and Small, respectively. Columns show (a) ground-truth depth, (b) final reconstruction obtained by combining reliable FPP depth with MMDC-Net predictions, (c) MMDC-Net prediction in unreliable regions only, (d) ground-truth depth restricted to the unreliable regions, and (e) error map over the unreliable regions. The figure shows that the Base backbone provides a favorable balance between reconstruction fidelity and computational efficiency, whereas the Small backbone introduces visibly larger reconstruction errors.}
\label{fig:mmdcnet_depth_anything_comparison}
\end{figure}

To mitigate these geometric sensor limitations, our logic-driven pipeline selectively triggers the MMDC-Net upon detecting that the HDD is Platter Facing, as explained in subsection \ref{sec:logic} . We use this to fuse the reliable FPP depth measurements with relative depth priors from the Depth Anything V2 as well as the projector illuminated grayscale image, and thereby accurately predict the depth of these unreliable regions, effectively resolving the artifacts. In this way, our hybrid approach yields a complete, high-fidelity 3D profile of the hard disk drive. 
The original network, however, utilizes Depth Anything V2 Large, with a resized input resolution of $1596 \times 1512$ for increased fidelity of the relative depth map. This comes at significant computational cost. Therefore, we explore more computationally efficient Depth Anything V2 backbones, utilizing a lower input resolution, with both the qualitative and quantitative results outlined in Fig. \ref{fig:mmdcnet_depth_anything_comparison} and Table \ref{tab:accuracy_results} respectively. We find from Table \ref{tab:accuracy_results} that our baseline model, while achieving the highest accuracy, runs at 0.4 Frames Per Second (FPS). Predictably, our fastest backbone is Depth Anything V2 Small, reaching a throughput of 76.2 FPS. However, this comes with a significant accuracy penalty, observed both qualitatively in Fig. \ref{fig:mmdcnet_depth_anything_comparison} and quantitatively in Table \ref{tab:accuracy_results}. In the same vein, we find that Depth Anything V2 Base, utilized with the default input resolution of $546 \times 518$, has an RMSE and MAE comparable to our Depth Anything V2 Large Backbone with High Resolution Input, while still resulting in a speedup from 0.4 FPS to 32.1 FPS. Due to this, we pick the Depth Anything V2 Base backbone for deployment in our final vision pipeline.

\subsection{Integrated 2D-3D Output with Pixel-Wise Correspondence}
\begin{figure}[pos=htbp]
\centering
\includegraphics[width=0.75\textwidth]{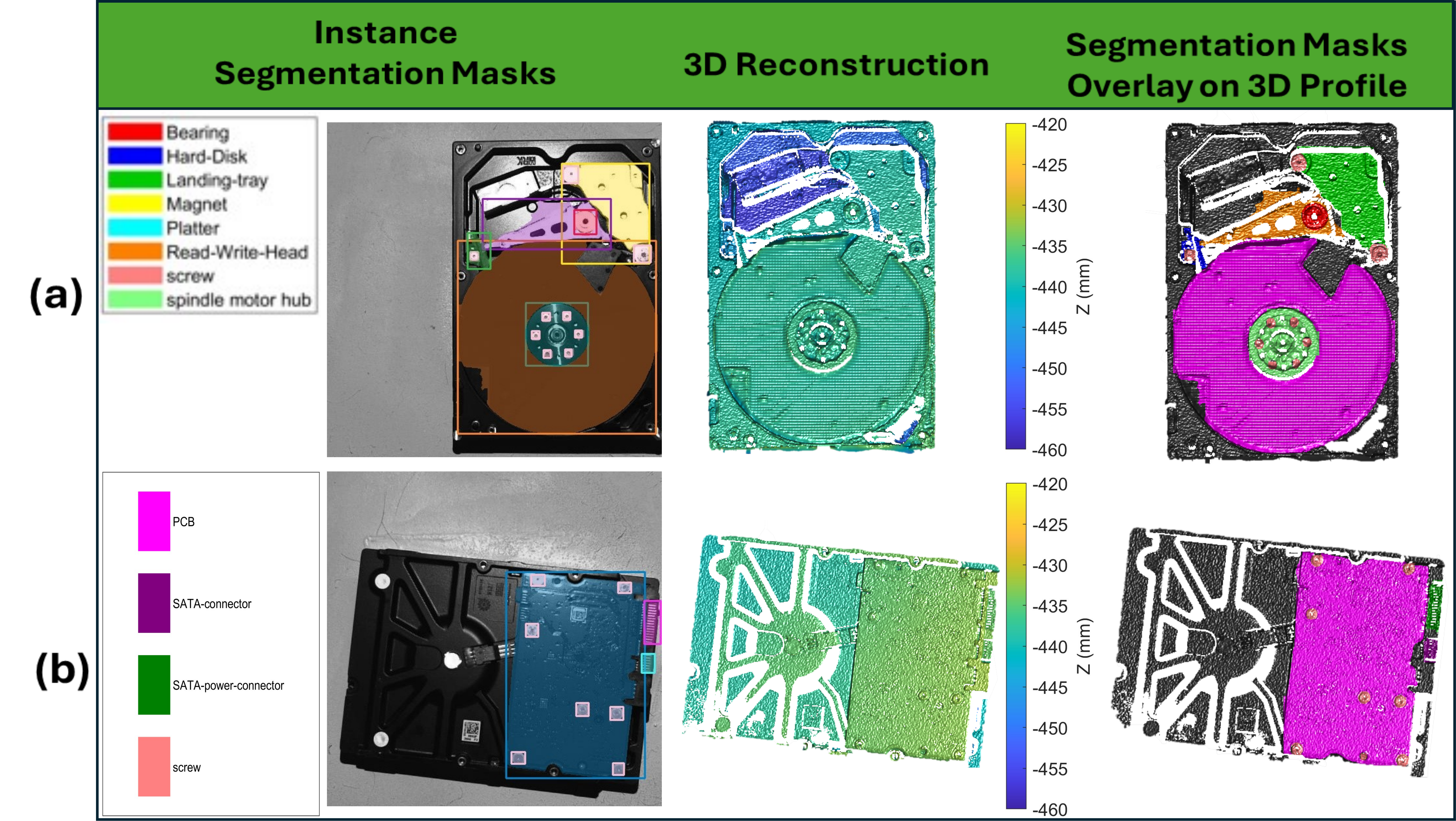}

\caption{Integrated semantic and geometric output of the proposed vision pipeline for representative Platter Facing and PCB Facing HDDs. Rows (a) and (b) correspond to the Platter Facing and PCB Facing cases, respectively. Within each row, the columns show: the predicted instance segmentation masks overlaid on the projector-illuminated input image, the reconstructed 3D profile obtained from the adaptive FPP pipeline, and the final semantic-geometry fusion obtained by projecting the 2D segmentation masks directly onto the 3D reconstruction. The figure illustrates that, because both the 2D semantic masks and the 3D depth map are derived from the same camera viewpoint, semantic labeling of the reconstructed HDD surface is obtained automatically without a separate registration step.}
\label{fig:combined_output}
\end{figure}

To demonstrate the comprehensive scene understanding capabilities of our pipeline, Fig. \ref{fig:combined_output} presents the synthesized output of the proposed system. By mapping the 2D instance segmentation masks directly onto the 3D point cloud, we illustrate the utility of our imaging approach for downstream disassembly planning. 

The instance segmentation masks provide the semantic hierarchy necessary for targeted path planning and disassembly. By evaluating whether predicted fastener masks fall within the convex hull of larger macro-components, the system can autonomously associate specific screws with their parent structures (e.g., isolating spindle screws from PCB screws). 

A primary advantage of our integrated architecture is the inherent pixel-wise correspondence provided by the FPP setup. Because the 2D projector-illuminated grayscale images (utilized by YOLOv11n for instance segmentation and MMDC-Net for depth completion) and the fringe-patterned images (utilized for 3D reconstruction) are captured by the same camera, the resulting semantic masks and computed depth map are inherently aligned in the image plane. This shared optical axis eliminates the need for complex multi-sensor extrinsic calibration, stereo registration, or subsequent point-cloud alignment, thereby streamlining the computational pipeline for real-time robotic manipulation. This provides an advantage over standard industrial RGB-D setups such as the Intel RealSense, Microsoft Azure Kinect, or paired 2D/3D camera rigs, whose color and depth sensing often originate from physically separate imaging elements. For example, unlike the robotic disassembly architecture proposed by Yildiz et al. \cite{yildiz_visual_2020}, which must continuously execute k-nearest-neighbor regression to map 2D semantic detections from a top-down camera onto a 3D point cloud generated by a physically offset stereo sensor, our shared-camera acquisition setup provides direct spatial correspondence between semantic masks and reconstructed geometry.

\begin{table}[htbp]
\centering
\caption{Quantitative evaluation of the MMDC-Net fusion module. The first row represents our original baseline configuration \cite{balasubramaniam_application-driven_2026}, while subsequent rows demonstrate the performance of the module when retrained with more efficient backbones. Metrics reported are Root Mean Square Error (RMSE), Mean Absolute Error (MAE), and original inference throughput (FPS).}
\label{tab:accuracy_results}
\resizebox{\columnwidth}{!}{%
\begin{tabular}{@{}llccccc@{}} 
\toprule
\multirow{2}{*}{\textbf{Backbone Used}} & \multirow{2}{*}{\textbf{\shortstack{Input\\Res.}}} & \multirow{2}{*}{\textbf{\shortstack{Orig.\\FPS}}} & \multicolumn{2}{c}{\textbf{Test Set (mm)}} & \multicolumn{2}{c}{\textbf{Validation Set (mm)}} \\ \cmidrule(lr){4-5} \cmidrule(lr){6-7}
 &  &  & \textbf{RMSE} & \textbf{MAE} & \textbf{RMSE} & \textbf{MAE} \\ \midrule
\multicolumn{7}{c}{\textbf{Method: MMDC-Net}} \\ \midrule
Depth Anything V2 Large (High-Res, Baseline \cite{balasubramaniam_application-driven_2026}) & $1596 \times 1512$ & 0.4 & 1.890 & 1.210 & 3.050 & 2.690 \\
Depth Anything V2 Large & $546 \times 518$ & 11.8 & 2.403 & 1.636 & 3.382 & 2.987 \\
\textbf{Depth Anything V2 Base }& \textbf{$546 \times 518$} & \textbf{32.1} & \textbf{2.317} & \textbf{1.836} & \textbf{3.291} & \textbf{2.754} \\
Depth Anything V2 Small & $546 \times 518$ & 76.2 & 4.848 & 4.563 & 4.579 & 3.965 \\ \bottomrule
\end{tabular}%
}
\end{table}
\subsection{Inference Optimization and Deployment Considerations}\label{sec:deployment}
To evaluate the practical runtime characteristics of the learned inference stack, we optimized memory usage and inference time on the evaluation workstation. We performed Half Precision (FP16) Optimization \cite{b_accelerating_2024} on MMDC-Net, Depth Anything V2, and YOLOv11n using the ONNX and TensorRT frameworks \cite{noauthor_introduction_nodate,noauthor_nvidia_nodate,zhou_exploring_2022}. This configuration has empirically demonstrated a good trade-off between latency and accuracy \cite{sumaiya_enhancing_2024}. We selected FP16 optimization over INT8 quantization to avoid the need for a calibration dataset, which is critical given our constraint of a small real-world training set. 

We present a comparison of inference time, and memory usage before and after the optimization of each of the neural networks in our pipeline in Table \ref{tab:deployment_metrics}. The throughput was averaged out over 1000 iterations after a warmup of 100 iterations. The results indicate a substantial reduction in memory requirements, ranging from 39\% for YOLOv11n to over 73\% for the MMDC-Net depth completion module. Furthermore, TensorRT yielded significant throughput gains, with a speedup of approximately 400\% - 850\% across the modules. As a reminder, our chosen Depth Anything V2 backbone, is the Depth Anything V2 Base Backbone, due to the speed and accuracy trade-offs outlined in section \ref{sec:results_hdd_faceup}. This yields a selectively triggered Platter Facing learned inference stack comprising YOLOv11n-seg, Depth Anything V2 Base, and MMDC-Net, with a combined neural inference latency of 12.86 ms and a throughput of 77.7 FPS on the evaluation workstation.

\begin{table}[htbp]
\centering
\caption{Inference performance of the proposed pipeline components on the evaluation workstation. The input resolution for the Depth Anything V2 backbone was varied to evaluate the trade-off between spatial precision and latency. Note that the final Depth Anything V2 output MMDC-Net fusion module operates at a fixed resolution of $512 \times 512$.}
\label{tab:deployment_metrics}
\resizebox{\columnwidth}{!}{%
\begin{tabular}{@{}lccccc@{}}
\toprule
\multirow{2}{*}{\textbf{Model}} & \multirow{2}{*}{\textbf{\shortstack{Resized Input\\Resolution}}} & \multicolumn{2}{c}{\textbf{Throughput (FPS)}} & \multicolumn{2}{c}{\textbf{GPU Memory (MB)}} \\ \cmidrule(lr){3-4} \cmidrule(lr){5-6}
 &  & \textbf{Original} & \textbf{Optimized} & \textbf{Original} & \textbf{Optimized} \\ \midrule
\multicolumn{6}{c}{\textit{Component: Segmentation \& Localization}} \\ \midrule
\textbf{YOLOv11n-seg} & $512 \times 512$ & 269.6 & \textbf{1593.0} & 180 & \textbf{110} \\ \midrule
\multicolumn{6}{c}{\textit{Component: Depth Estimation Backbones}} \\ \midrule
Depth Anything V2 (Small) & $546 \times 518$ & 76.2 & 654.4 & 380 & 180 \\
\textbf{Depth Anything V2 (Base)} & $546 \times 518$ & 32.1 & \textbf{277.4} & 814 & \textbf{334} \\
Depth Anything V2 (Large) & $546 \times 518$ & 11.8 & 98.4 & 1924 & 892 \\
Depth Anything V2 (Large-HighRes) & $1596 \times 1512$ & 0.4 & 5.0 & 29448 & 1990 \\ \midrule
\multicolumn{6}{c}{\textit{Component: Depth Completion Module}} \\ \midrule
\textbf{MMDC-Net} & $512 \times 512$ & 23.2 & \textbf{115.8} & 5440 & \textbf{1454} \\ \bottomrule
\end{tabular}%
}
\end{table}

\section{Limitations}\label{sec12}
While the proposed autonomous vision pipeline significantly advances the capabilities of robotic e-waste disassembly, we have a few limitations.

Our baseline FPP system is inherently susceptible to occlusion-induced shadows. Since FPP relies on the unoccluded intersection of the camera's field of view and the projector's illumination path, sudden depth discontinuities, such as tall internal components or deep cavities within the drive casing, can block the projected fringe patterns. Our current framework does not algorithmically fill these resultant data voids unless they occur on the selectively triggered, reflective platter.

Additionally, our synthetic digital twin and the resulting open-sourced instance segmentation dataset are intrinsically tied to a specific geometric and optical calibration configuration. Models pre-trained on this dataset may experience performance degradation if deployed out-of-the-box on a physical FPP system possessing substantially different intrinsic or extrinsic parameters (e.g., a different camera-projector baseline or focal length). Nevertheless, this limitation can be effectively mitigated through fine-tuning on a small sample of target-domain data. While matching the digital twin's calibration to the target real-world system is optimal, fine-tuning can be highly viable even with a calibration mismatch, as 2D instance segmentation relies on semantic and structural features and is fundamentally less sensitive to strict geometric calibration than pixel-wise 3D depth prediction tasks.
\section{Conclusions and Future Work}\label{sec13}
This work presented an autonomous vision pipeline for robotic hard disk drive disassembly that bridges 2D semantic scene understanding and high-fidelity 3D metrology. By integrating an instance segmentation model with FPP, the system not only localizes critical components and fasteners but also dictates a logic-driven 3D scanning protocol. Specifically, detected platter presence is used to selectively trigger our depth completion module, MMDC-Net. Furthermore, we demonstrated a sim-to-real workflow, while performing throughput optimization using TensorRT to support practical low-latency inference and inform future deployment on target hardware. Quantitatively, the proposed system achieved a low latency suitable for resource-constrained real-time inference settings, with Box mAP@50 of $0.960$, Box mAP@50--95 of $0.809$, Mask mAP@50 of $0.957$, and Mask mAP@50--95 of $0.724$ for instance segmentation, while the selected MMDC-Net configuration with the Depth Anything V2 Base backbone achieved a test RMSE of \(2.317\) mm and MAE of \(1.836\) mm, and the selectively triggered Platter Facing learned inference stack achieved a combined inference latency of \(12.86\) ms and a throughput of \(77.7\) FPS on the evaluation workstation.

Future work will focus on integrating this vision pipeline directly with a robotic manipulator to execute closed-loop, physical disassembly sequences based on the generated semantic and spatial data. Additionally, we plan to expand the synthetic data generation and logic-driven scanning frameworks to encompass a broader taxonomy of complex electronic waste, further advancing scalable material recovery efforts.

% \appendix
% \section{Section title of first appendix}\label{secA1}

% \bibliographystyle{cas-model2-names}
% \FloatBarrier
% \clearpage
\bibliographystyle{elsarticle-num}
\bibliography{sn-bibliography}

\begin{thebibliography}{10}
\expandafter\ifx\csname url\endcsname\relax
  \def\url#1{\texttt{#1}}\fi
\expandafter\ifx\csname urlprefix\endcsname\relax\def\urlprefix{URL }\fi
\expandafter\ifx\csname href\endcsname\relax
  \def\href#1#2{#2} \def\path#1{#1}\fi

\bibitem{balde2024global}
C.~P. Bald\'{e}, R.~Kuehr, T.~Yamamoto, R.~McDonald, E.~D’Angelo, S.~Althaf, G.~Bel, O.~Deubzer, E.~Fernandez-Cubillo, V.~Forti, V.~Gray, S.~Herat, S.~Honda, G.~Iattoni, D.~S. Khetriwal, V.~Luda~di Cortemiglia, Y.~Lobuntsova, I.~Nnorom, N.~Pralat, M.~Wagner, The global e-waste monitor 2024, Tech. rep., United Nations Institute for Training and Research (UNITAR) and International Telecommunication Union (ITU), Geneva/Bonn, cC BY-NC-SA 3.0 IGO (2024).

\bibitem{sabbaghi_global_2019}
M.~Sabbaghi, W.~Cade, W.~Olson, S.~Behdad, \href{https://onlinelibrary.wiley.com/doi/abs/10.1111/jiec.12765}{The {Global} {Flow} of {Hard} {Disk} {Drives}: {Quantifying} the {Concept} of {Value} {Leakage} in {E}-waste {Recovery} {Systems}}, Journal of Industrial Ecology 23~(3) (2019) 560--573, \_eprint: https://onlinelibrary.wiley.com/doi/pdf/10.1111/jiec.12765.
\newblock \href {https://doi.org/10.1111/jiec.12765} {\path{doi:10.1111/jiec.12765}}.
\newline\urlprefix\url{https://onlinelibrary.wiley.com/doi/abs/10.1111/jiec.12765}

\bibitem{habib_tracking_2015}
K.~Habib, K.~Parajuly, H.~Wenzel, \href{https://doi.org/10.1021/acs.est.5b02264}{Tracking the {Flow} of {Resources} in {Electronic} {Waste} - {The} {Case} of {End}-of-{Life} {Computer} {Hard} {Disk} {Drives}}, Environmental Science \& Technology 49~(20) (2015) 12441--12449.
\newblock \href {https://doi.org/10.1021/acs.est.5b02264} {\path{doi:10.1021/acs.est.5b02264}}.
\newline\urlprefix\url{https://doi.org/10.1021/acs.est.5b02264}

\bibitem{tanvar_characterization_2020}
H.~Tanvar, A.~Barnwal, N.~Dhawan, \href{https://www.sciencedirect.com/science/article/pii/S0959652619342477}{Characterization and evaluation of discarded hard disc drives for recovery of copper and rare earth values}, Journal of Cleaner Production 249 (2020) 119377.
\newblock \href {https://doi.org/10.1016/j.jclepro.2019.119377} {\path{doi:10.1016/j.jclepro.2019.119377}}.
\newline\urlprefix\url{https://www.sciencedirect.com/science/article/pii/S0959652619342477}

\bibitem{liu_global_2023}
K.~Liu, Q.~Tan, J.~Yu, M.~Wang, \href{https://www.sciencedirect.com/science/article/pii/S2773167723000055}{A global perspective on e-waste recycling}, Circular Economy 2~(1) (2023) 100028.
\newblock \href {https://doi.org/10.1016/j.cec.2023.100028} {\path{doi:10.1016/j.cec.2023.100028}}.
\newline\urlprefix\url{https://www.sciencedirect.com/science/article/pii/S2773167723000055}

\bibitem{barman_optimizing_2025}
T.~Barman, S.~Coleman, D.~Kerr, S.~Harrigan, J.~Quinn, \href{https://ieeexplore.ieee.org/document/10995066}{Optimizing {Industrial} {E}-{Waste} {Recycling} with {Attention}-{Driven} {Deep} {Learning} for {PCB} {Segmentation} {Using} {Hyperspectral} {Imaging}}, in: 2025 {IEEE} {Symposia} on {Computational} {Intelligence} for {Energy}, {Transport} and {Environmental} {Sustainability} ({CIETES}), 2025, pp. 1--7.
\newblock \href {https://doi.org/10.1109/CIETES63869.2025.10995066} {\path{doi:10.1109/CIETES63869.2025.10995066}}.
\newline\urlprefix\url{https://ieeexplore.ieee.org/document/10995066}

\bibitem{hussein_harnessing_2025}
A.~F. Hussein, W.~M. Hamanah, M.~A. Abido, \href{https://www.sciencedirect.com/science/article/pii/S2590123025021826}{Harnessing hyperspectral imaging and deep learning for advanced e-waste classification using three spectral bands}, Results in Engineering 27 (2025) 106110.
\newblock \href {https://doi.org/10.1016/j.rineng.2025.106110} {\path{doi:10.1016/j.rineng.2025.106110}}.
\newline\urlprefix\url{https://www.sciencedirect.com/science/article/pii/S2590123025021826}

\bibitem{picon_analysis_2025}
A.~Picon, P.~Galan, A.~Bereciartua-Perez, L.~Benito-del Valle, \href{https://www.sciencedirect.com/science/article/pii/S1386142524018316}{On the analysis of adapting deep learning methods to hyperspectral imaging. {Use} case for {WEEE} recycling and dataset}, Spectrochimica Acta Part A: Molecular and Biomolecular Spectroscopy 330 (2025) 125665.
\newblock \href {https://doi.org/10.1016/j.saa.2024.125665} {\path{doi:10.1016/j.saa.2024.125665}}.
\newline\urlprefix\url{https://www.sciencedirect.com/science/article/pii/S1386142524018316}

\bibitem{mohsin_heatmap_2025}
M.~Mohsin, S.~Rovetta, F.~Masulli, A.~Cabri, \href{https://ieeexplore.ieee.org/document/11086768}{Heatmap {Visualization} for {Deep} {Learning} {Analysis} of {Waste} {Printed} {Circuit} {Boards}}, in: 2025 {IEEE} 2nd {International} {Conference} on {Electronics}, {Communications} and {Intelligent} {Science} ({ECIS}), 2025, pp. 1--5.
\newblock \href {https://doi.org/10.1109/ECIS65594.2025.11086768} {\path{doi:10.1109/ECIS65594.2025.11086768}}.
\newline\urlprefix\url{https://ieeexplore.ieee.org/document/11086768}

\bibitem{selvaraju_grad-cam_2017}
R.~R. Selvaraju, M.~Cogswell, A.~Das, R.~Vedantam, D.~Parikh, D.~Batra, \href{https://ieeexplore.ieee.org/document/8237336}{Grad-{CAM}: {Visual} {Explanations} from {Deep} {Networks} via {Gradient}-{Based} {Localization}}, in: 2017 {IEEE} {International} {Conference} on {Computer} {Vision} ({ICCV}), 2017, pp. 618--626, iSSN: 2380-7504.
\newblock \href {https://doi.org/10.1109/ICCV.2017.74} {\path{doi:10.1109/ICCV.2017.74}}.
\newline\urlprefix\url{https://ieeexplore.ieee.org/document/8237336}

\bibitem{ahmed_exploration_2025}
S.~Ahmed, D.~F. Sanam, \href{https://ieeexplore.ieee.org/abstract/document/11160027}{Exploration of an {E}-waste {Prediction} {Model} for {Collection} {System} {Optimization} {Using} {Deep} {Learning} {Models} in {E}-waste {Management} {Facilities}}, in: 2025 2nd {International} {Conference} on {Next}-{Generation} {Computing}, {IoT} and {Machine} {Learning} ({NCIM}), 2025, pp. 1--6.
\newblock \href {https://doi.org/10.1109/NCIM65934.2025.11160027} {\path{doi:10.1109/NCIM65934.2025.11160027}}.
\newline\urlprefix\url{https://ieeexplore.ieee.org/abstract/document/11160027}

\bibitem{jahanian_see_2019}
A.~Jahanian, Q.~H. Le, K.~Youcef-Toumi, D.~Tsetserukou, \href{https://openaccess.thecvf.com/content_CVPRW_2019/html/cv4gc/Jahanian_See_the_E-Waste_Training_Visual_Intelligence_to_See_Dense_Circuit_CVPRW_2019_paper.html}{See the {E}-{Waste}! {Training} {Visual} {Intelligence} to {See} {Dense} {Circuit} {Boards} for {Recycling}}, 2019, pp. 0--0.
\newline\urlprefix\url{https://openaccess.thecvf.com/content_CVPRW_2019/html/cv4gc/Jahanian_See_the_E-Waste_Training_Visual_Intelligence_to_See_Dense_Circuit_CVPRW_2019_paper.html}

\bibitem{liu_raise_2026}
C.~Liu, B.~Balasubramaniam, N.~A. Yancey, M.~H. Severson, A.~Shine, P.~Bove, B.~Li, X.~Liang, M.~Zheng, \href{https://www.sciencedirect.com/science/article/pii/S0921344925004860}{{RAISE}: {A} {Robot}-{Assisted} {SelectIve} {Disassembly} and {Sorting} {System} for {End}-of-{Life} {Phones}}, Resources, Conservation and Recycling 225 (2026) 108609.
\newblock \href {https://doi.org/10.1016/j.resconrec.2025.108609} {\path{doi:10.1016/j.resconrec.2025.108609}}.
\newline\urlprefix\url{https://www.sciencedirect.com/science/article/pii/S0921344925004860}

\bibitem{foo_screw_2021}
G.~Foo, S.~Kara, M.~Pagnucco, \href{https://www.sciencedirect.com/science/article/pii/S2212827121002031}{Screw detection for disassembly of electronic waste using reasoning and re-training of a deep learning model}, Procedia CIRP 98 (2021) 666--671.
\newblock \href {https://doi.org/10.1016/j.procir.2021.01.172} {\path{doi:10.1016/j.procir.2021.01.172}}.
\newline\urlprefix\url{https://www.sciencedirect.com/science/article/pii/S2212827121002031}

\bibitem{karbouj_comparative_2024}
B.~Karbouj, G.~A. Topalian-Rivas, J.~Krüger, \href{https://www.sciencedirect.com/science/article/pii/S2212827124001021}{Comparative {Performance} {Evaluation} of {One}-{Stage} and {Two}-{Stage} {Object} {Detectors} for {Screw} {Head} {Detection} and {Classification} in {Disassembly} {Processes}}, Procedia CIRP 122 (2024) 527--532.
\newblock \href {https://doi.org/10.1016/j.procir.2024.01.077} {\path{doi:10.1016/j.procir.2024.01.077}}.
\newline\urlprefix\url{https://www.sciencedirect.com/science/article/pii/S2212827124001021}

\bibitem{puttero_automatic_2024}
S.~Puttero, A.~Nassehi, E.~Verna, G.~Genta, M.~Galetto, \href{https://www.sciencedirect.com/science/article/pii/S2212827124003457}{Automatic object detection for disassembly and recycling of electronic board components}, Procedia CIRP 127 (2024) 206--211.
\newblock \href {https://doi.org/10.1016/j.procir.2024.07.036} {\path{doi:10.1016/j.procir.2024.07.036}}.
\newline\urlprefix\url{https://www.sciencedirect.com/science/article/pii/S2212827124003457}

\bibitem{brogan_deep_2021}
D.~P. Brogan, N.~M. DiFilippo, M.~K. Jouaneh, \href{https://www.sciencedirect.com/science/article/pii/S2590005621000394}{Deep learning computer vision for robotic disassembly and servicing applications}, Array 12 (2021) 100094.
\newblock \href {https://doi.org/10.1016/j.array.2021.100094} {\path{doi:10.1016/j.array.2021.100094}}.
\newline\urlprefix\url{https://www.sciencedirect.com/science/article/pii/S2590005621000394}

\bibitem{zhao_precision_2025}
H.~Zhao, C.~Liu, B.~Balasubramaniam, J.~Li, J.~Song, X.~Liang, M.~Zheng, B.~Li, \href{https://opg.optica.org/ao/abstract.cfm?uri=ao-64-30-8986}{Precision {3D} profilometry of consumer-grade computer enclosures using high dynamic range fringe projection}, Applied Optics 64~(30) (2025) 8986--8994.
\newblock \href {https://doi.org/10.1364/AO.575966} {\path{doi:10.1364/AO.575966}}.
\newline\urlprefix\url{https://opg.optica.org/ao/abstract.cfm?uri=ao-64-30-8986}

\bibitem{yildiz_visual_2020}
E.~Yildiz, T.~Brinker, E.~Renaudo, J.~Hollenstein, S.~Haller-Seeber, J.~Piater, F.~Wörgötter, \href{https://www.scitepress.org/DigitalLibrary/Link.aspx?doi=10.5220/0010016000170027}{A {Visual} {Intelligence} {Scheme} for {Hard} {Drive} {Disassembly} in {Automated} {Recycling} {Routines}:}, in: Proceedings of the {International} {Conference} on {Robotics}, {Computer} {Vision} and {Intelligent} {Systems}, SCITEPRESS - Science and Technology Publications, Budapest, Hungary, 2020, pp. 17--27.
\newblock \href {https://doi.org/10.5220/0010016000170027} {\path{doi:10.5220/0010016000170027}}.
\newline\urlprefix\url{https://www.scitepress.org/DigitalLibrary/Link.aspx?doi=10.5220/0010016000170027}

\bibitem{yildiz_extended_2022}
E.~Yildiz, E.~Renaudo, J.~Hollenstein, J.~Piater, F.~Wörgötter, An {Extended} {Visual} {Intelligence} {Scheme} for {Disassembly} in {Automated} {Recycling} {Routines}, in: P.~Galambos, E.~Kayacan, K.~Madani (Eds.), Robotics, {Computer} {Vision} and {Intelligent} {Systems}, Springer International Publishing, Cham, 2022, pp. 25--50.
\newblock \href {https://doi.org/10.1007/978-3-031-19650-8_2} {\path{doi:10.1007/978-3-031-19650-8_2}}.

\bibitem{yildiz_dcnn-based_2019}
E.~Yildiz, F.~Wörgötter, \href{https://ieeexplore.ieee.org/document/9067965}{{DCNN}-{Based} {Screw} {Detection} for {Automated} {Disassembly} {Processes}}, in: 2019 15th {International} {Conference} on {Signal}-{Image} {Technology} \& {Internet}-{Based} {Systems} ({SITIS}), 2019, pp. 187--192.
\newblock \href {https://doi.org/10.1109/SITIS.2019.00040} {\path{doi:10.1109/SITIS.2019.00040}}.
\newline\urlprefix\url{https://ieeexplore.ieee.org/document/9067965}

\bibitem{rojas_efficient_2023}
C.~Rojas, A.~Rodríguez-Sánchez, E.~Renaudo, \href{https://www.worldscientific.com/doi/abs/10.1142/9789811289125_0005}{Efficient {Segmentation} of {E}-{Waste} {Devices} {With} {Deep} {Learning} for {Robotic} {Recycling}}, in: Emerging {Topics} in {Pattern} {Recognition} and {Artificial} {Intelligence}, Vol. Volume 9 of Series on {Language} {Processing}, {Pattern} {Recognition}, and {Intelligent} {Systems}, WORLD SCIENTIFIC, 2023, pp. 123--144.
\newblock \href {https://doi.org/10.1142/9789811289125_0005} {\path{doi:10.1142/9789811289125_0005}}.
\newline\urlprefix\url{https://www.worldscientific.com/doi/abs/10.1142/9789811289125_0005}

\bibitem{rojas_deep_2022}
C.~Rojas, A.~Rodríguez-Sánchez, E.~Renaudo, Deep {Learning} for {Fast} {Segmentation} of {E}-waste {Devices}’ {Inner} {Parts} in a {Recycling} {Scenario}, in: M.~El~Yacoubi, E.~Granger, P.~C. Yuen, U.~Pal, N.~Vincent (Eds.), Pattern {Recognition} and {Artificial} {Intelligence}, Springer International Publishing, Cham, 2022, pp. 161--172.
\newblock \href {https://doi.org/10.1007/978-3-031-09037-0_14} {\path{doi:10.1007/978-3-031-09037-0_14}}.

\bibitem{li_high-speed_2022}
B.~Li, \href{https://revistas.utb.edu.co/tesea/article/view/490}{High-speed {3D} optical sensing for manufacturing research and industrial sensing applications}, Transactions on Energy Systems and Engineering Applications 3~(2) (2022) 1--12.
\newblock \href {https://doi.org/10.32397/tesea.vol3.n2.490} {\path{doi:10.32397/tesea.vol3.n2.490}}.
\newline\urlprefix\url{https://revistas.utb.edu.co/tesea/article/view/490}

\bibitem{balasubramaniam_application-driven_2026}
B.~Balasubramaniam, V.~Suresh, Y.~Cheng, J.~Li, B.~Li, \href{https://www.sciencedirect.com/science/article/pii/S0143816625007717}{Application-driven multi-modal depth completion in fringe projection profilometry}, Optics and Lasers in Engineering 200 (2026) 109587.
\newblock \href {https://doi.org/10.1016/j.optlaseng.2025.109587} {\path{doi:10.1016/j.optlaseng.2025.109587}}.
\newline\urlprefix\url{https://www.sciencedirect.com/science/article/pii/S0143816625007717}

\bibitem{noauthor_imagine_nodate}
\href{https://imagine-h2020.eu/hdd-taxonomy.php}{{IMAGINE}}.
\newline\urlprefix\url{https://imagine-h2020.eu/hdd-taxonomy.php}

\bibitem{khanam_yolov11_2024}
R.~Khanam, M.~Hussain, \href{http://arxiv.org/abs/2410.17725}{{YOLOv11}: {An} {Overview} of the {Key} {Architectural} {Enhancements}}, arXiv:2410.17725 [cs] version: 1 (Oct. 2024).
\newblock \href {https://doi.org/10.48550/arXiv.2410.17725} {\path{doi:10.48550/arXiv.2410.17725}}.
\newline\urlprefix\url{http://arxiv.org/abs/2410.17725}

\bibitem{wang_non-contact_2025}
Q.~Wang, X.~Wei, K.~Li, B.~Cao, W.~Zhang, \href{https://www.mdpi.com/2077-0472/15/21/2180}{Non-{Contact} {Measurement} of {Sunflower} {Flowerhead} {Morphology} {Using} {Mobile}-{Boosted} {Lightweight} {Asymmetric} ({MBLA})-{YOLO} and {Point} {Cloud} {Technology}}, Agriculture 15~(21) (2025) 2180.
\newblock \href {https://doi.org/10.3390/agriculture15212180} {\path{doi:10.3390/agriculture15212180}}.
\newline\urlprefix\url{https://www.mdpi.com/2077-0472/15/21/2180}

\bibitem{naqvi_four-dimensional_2025}
L.~H. Naqvi, B.~Balasubramaniam, J.~Li, L.~Liu, B.~Li, \href{https://www.mdpi.com/2077-0472/15/15/1702}{Four-{Dimensional} {Hyperspectral} {Imaging} for {Fruit} and {Vegetable} {Grading}}, Agriculture 15~(15) (2025) 1702.
\newblock \href {https://doi.org/10.3390/agriculture15151702} {\path{doi:10.3390/agriculture15151702}}.
\newline\urlprefix\url{https://www.mdpi.com/2077-0472/15/15/1702}

\bibitem{zhang_high-speed_2018}
S.~Zhang, High-{Speed} {3D} {Imaging} with {Digital} {Fringe} {Projection} {Techniques}, CRC Press, 2018, google-Books-ID: p22mCwAAQBAJ.

\bibitem{li_novel_2014}
B.~Li, N.~Karpinsky, S.~Zhang, \href{https://opg.optica.org/ao/abstract.cfm?uri=ao-53-16-3415}{Novel calibration method for structured-light system with an out-of-focus projector}, Applied Optics 53~(16) (2014) 3415--3426, publisher: Optica Publishing Group.
\newblock \href {https://doi.org/10.1364/AO.53.003415} {\path{doi:10.1364/AO.53.003415}}.
\newline\urlprefix\url{https://opg.optica.org/ao/abstract.cfm?uri=ao-53-16-3415}

\bibitem{balasubramaniam_3d_2023}
B.~Balasubramaniam, J.~Li, L.~Liu, B.~Li, \href{https://www.mdpi.com/2079-9292/12/4/859}{{3D} {Imaging} with {Fringe} {Projection} for {Food} and {Agricultural} {Applications}—{A} {Tutorial}}, Electronics 12~(4) (2023) 859, number: 4.
\newblock \href {https://doi.org/10.3390/electronics12040859} {\path{doi:10.3390/electronics12040859}}.
\newline\urlprefix\url{https://www.mdpi.com/2079-9292/12/4/859}

\bibitem{balasubramaniam_single_2023}
B.~Balasubramaniam, B.~Li, \href{https://doi.org/10.1115/MSEC2023-104380}{Single {Shot} {3D} {Shape} {Measurement} of {Non}-{Volatile} {Data} {Storage} {Devices}}, Vol. Volume 2: Manufacturing Equipment and Automation; Manufacturing Processes; Manufacturing Systems; Nano/Micro/Meso Manufacturing; Quality and Reliability of International {Manufacturing} {Science} and {Engineering} {Conference}, 2023, \_eprint: https://asmedigitalcollection.asme.org/MSEC/proceedings-pdf/MSEC2023/87240/V002T06A010/7046555/v002t06a010-msec2023-104380.pdf.
\newblock \href {https://doi.org/10.1115/MSEC2023-104380} {\path{doi:10.1115/MSEC2023-104380}}.
\newline\urlprefix\url{https://doi.org/10.1115/MSEC2023-104380}

\bibitem{roboflow2026}
B.~Dwyer, J.~Nelson, T.~Hansen, et~al., \href{https://roboflow.com}{Roboflow ({Version} 1.0) [{Software}]}, computer vision (2026).
\newline\urlprefix\url{https://roboflow.com}

\bibitem{zheng_fringe_2020}
Y.~Zheng, S.~Wang, Q.~Li, B.~Li, \href{https://opg.optica.org/oe/abstract.cfm?uri=oe-28-24-36568}{Fringe projection profilometry by conducting deep learning from its digital twin}, Optics Express 28~(24) (2020) 36568--36583.
\newblock \href {https://doi.org/10.1364/OE.410428} {\path{doi:10.1364/OE.410428}}.
\newline\urlprefix\url{https://opg.optica.org/oe/abstract.cfm?uri=oe-28-24-36568}

\bibitem{noauthor_sketchfab_nodate}
\href{https://sketchfab.com}{Sketchfab - {The} best {3D} viewer on the web}.
\newline\urlprefix\url{https://sketchfab.com}

\bibitem{noauthor_grabcad_nodate}
\href{https://grabcad.com/}{{GrabCAD} {Making} {Additive} {Manufacturing} at {Scale} {Possible}}.
\newline\urlprefix\url{https://grabcad.com/}

\bibitem{carion2025sam}
N.~Carion, L.~Gustafson, Y.-T. Hu, S.~Debnath, R.~Hu, D.~Suris, C.~Ryali, K.~V. Alwala, H.~Khedr, A.~Huang, et~al., Sam 3: Segment anything with concepts, arXiv preprint arXiv:2511.16719 (2025).

\bibitem{lin_uninext_2023}
F.~Lin, J.~Yuan, S.~Wu, F.~Wang, Z.~Wang, \href{https://dl.acm.org/doi/10.1145/3581783.3612260}{{UniNeXt}: {Exploring} {A} {Unified} {Architecture} for {Vision} {Recognition}}, in: Proceedings of the 31st {ACM} {International} {Conference} on {Multimedia}, {MM} '23, Association for Computing Machinery, New York, NY, USA, 2023, pp. 3200--3208.
\newblock \href {https://doi.org/10.1145/3581783.3612260} {\path{doi:10.1145/3581783.3612260}}.
\newline\urlprefix\url{https://dl.acm.org/doi/10.1145/3581783.3612260}

\bibitem{b_accelerating_2024}
G.~P. B, R.~Marion Lincy~G, A.~Rishekeeshan, {Deekshitha}, \href{https://ieeexplore.ieee.org/abstract/document/10690032}{Accelerating {Native} {Inference} {Model} {Performance} in {Edge} {Devices} using {TensorRT}}, in: 2024 {IEEE} {Recent} {Advances} in {Intelligent} {Computational} {Systems} ({RAICS}), 2024, pp. 1--7, iSSN: 2769-5565.
\newblock \href {https://doi.org/10.1109/RAICS61201.2024.10690032} {\path{doi:10.1109/RAICS61201.2024.10690032}}.
\newline\urlprefix\url{https://ieeexplore.ieee.org/abstract/document/10690032}

\bibitem{noauthor_introduction_nodate}
\href{https://onnx.ai/onnx/intro/}{Introduction to {ONNX} - {ONNX} 1.21.0 documentation}.
\newline\urlprefix\url{https://onnx.ai/onnx/intro/}

\bibitem{noauthor_nvidia_nodate}
\href{https://developer.nvidia.com/tensorrt}{{NVIDIA} {TensorRT}}.
\newline\urlprefix\url{https://developer.nvidia.com/tensorrt}

\bibitem{zhou_exploring_2022}
Y.~Zhou, K.~Yang, \href{https://ieeexplore.ieee.org/abstract/document/10074837}{Exploring {TensorRT} to {Improve} {Real}-{Time} {Inference} for {Deep} {Learning}}, in: 2022 {IEEE} 24th {Int} {Conf} on {High} {Performance} {Computing} \& {Communications}; 8th {Int} {Conf} on {Data} {Science} \& {Systems}; 20th {Int} {Conf} on {Smart} {City}; 8th {Int} {Conf} on {Dependability} in {Sensor}, {Cloud} \& {Big} {Data} {Systems} \& {Application} ({HPCC}/{DSS}/{SmartCity}/{DependSys}), 2022, pp. 2011--2018.
\newblock \href {https://doi.org/10.1109/HPCC-DSS-SmartCity-DependSys57074.2022.00299} {\path{doi:10.1109/HPCC-DSS-SmartCity-DependSys57074.2022.00299}}.
\newline\urlprefix\url{https://ieeexplore.ieee.org/abstract/document/10074837}

\bibitem{sumaiya_enhancing_2024}
{Sumaiya}, R.~Jafarpourmarzouni, S.~Lu, Z.~Dong, \href{https://ieeexplore.ieee.org/abstract/document/10607059}{Enhancing {Real}-time {Inference} {Performance} for {Time}-{Critical} {Software}-{Defined} {Vehicles}}, in: 2024 {IEEE} {International} {Conference} on {Mobility}, {Operations}, {Services} and {Technologies} ({MOST}), 2024, pp. 101--113.
\newblock \href {https://doi.org/10.1109/MOST60774.2024.00019} {\path{doi:10.1109/MOST60774.2024.00019}}.
\newline\urlprefix\url{https://ieeexplore.ieee.org/abstract/document/10607059}

\end{thebibliography}
\end{document}